\pdfoutput=1
\documentclass[11pt]{article}

\usepackage{acl}

\usepackage{times}
\usepackage{latexsym}

\usepackage{multicol,multirow}

\usepackage[T1]{fontenc}
\usepackage[utf8]{inputenc}

\usepackage{microtype}

\usepackage{inconsolata}

\usepackage[shortlabels,inline]{enumitem}
\usepackage[normalem]{ulem}
\usepackage{graphicx}
\usepackage{todonotes}
\usepackage{caption}
\usepackage{subcaption}

\usepackage[type=none,unit=in]{fgruler}

\newcommand{\movedtoappendix}[1]{}
\usepackage[most]{tcolorbox}
\usepackage{cleveref}
\crefname{section}{s}{ss}
\crefname{section}{s}{ss}
\crefname{table}{Table}{Tables}
\crefname{figure}{Fig.}{Figs.}
\crefname{algorithm}{Alg.}{}
\crefname{ALC@unique}{Line}{Lines}
\crefname{equation}{Eq.}{Eqns.}
\crefname{appendix}{Appendix}{}
\crefformat{section}{\S#2#1#3}
\NewDocumentCommand{\prompt}{O{} +m}{%
\begin{tcolorbox}[
    coltitle=white,
    colframe=black,
    colback=black!5!white,
    boxrule=1pt,
    enhanced jigsaw,
    breakable,
    pad at break*=2mm,
    left=2pt,
    right=2pt,
    top=2pt,
    bottom=2pt,
    fontupper=\small,
    fontlower=\small,
    title={#1}, % Optional title
]
#2 % Multiline text content
\end{tcolorbox}
}
\newcommand{\promptvar}[1]{
    {\color{cyan}\{\{#1\}\}}
}

\title{SMART: State-based Multi Agent Reasoning Traces for Verification-oriented Problem-Solving}  

\author{ %
\textbf{Sai Vallurupalli, Francis Ferraro} \\
  Department of Computer Science and Electrical Engineering\\
  University of Maryland Baltimore County\\
  Baltimore, MD 21250 USA \\
  \texttt{\{kolli,ferraro\}@umbc.edu} \\
  }

\begin{document}
\maketitle

\begin{abstract}
\label{sec:abstract}

%Latent condition inferences underlying a complex problem can lead to new knowledge that can augment the problem context and useful for identifying a robust solution. However, knowing which latent of these conditional inferences are solution salient is challenging.  We propose a structured role-based prompting framework, LaCE, to obtain latent conditions that capture the implicit state of entities participating in the problem context.  From these curate a solution salient set of conditions and use it to augment the problem context.  Using 951 scientific and 432 non-scientific claims spanning 5 datasets as our complex problems, and a variety of Large Language Models, we show improvement in claim validation performance by up to 8\% where smaller models performance reachs the performance of larger models.  

Complex problems with limited context, such as claims, require a means to obtain additional knowledge necessary for solution reasoning. We propose a multi-agent framework, SMART, to obtain knowledge through capturing the implicit state of participating entities in a problem context.  SMART curates the state knowledge as diverse, complementary and solution-salient premises using a variety of LLMs and a structured role-based prompt strategy. Using knowledge from these real-world verified curated premises, SMART's knowledge guided solution reasoning improves validation performance up to 20\% on complex scientific claims spanning 3 datasets.   We show that grounding premises in real-world knowledge is necessary and improves performance up to 5\%.   
\end{abstract}

\section{Introduction}
\label{sec:intro}

%Validating a scientific claim is heavily dependent on tacit premises
%%Complex problem contexts do not contain all the information necessary for solution reasoning. This necessitates finding  crucial, likely latent or unspecified, aspects about relevant concepts, topics, or entities to reason and plan the solution~\cite{knauff_wolf,funke}..} 

Consider the problem of assessing the veracity of the following scientific claim,

\begin{quote}
\textit{Genetic deletion of JAM-A decreases gut permeability.}
\end{quote}

\begin{figure*}[t]
    \centering
    \includegraphics[trim=0 0in 0 0.1in,  width=1 \linewidth]
    {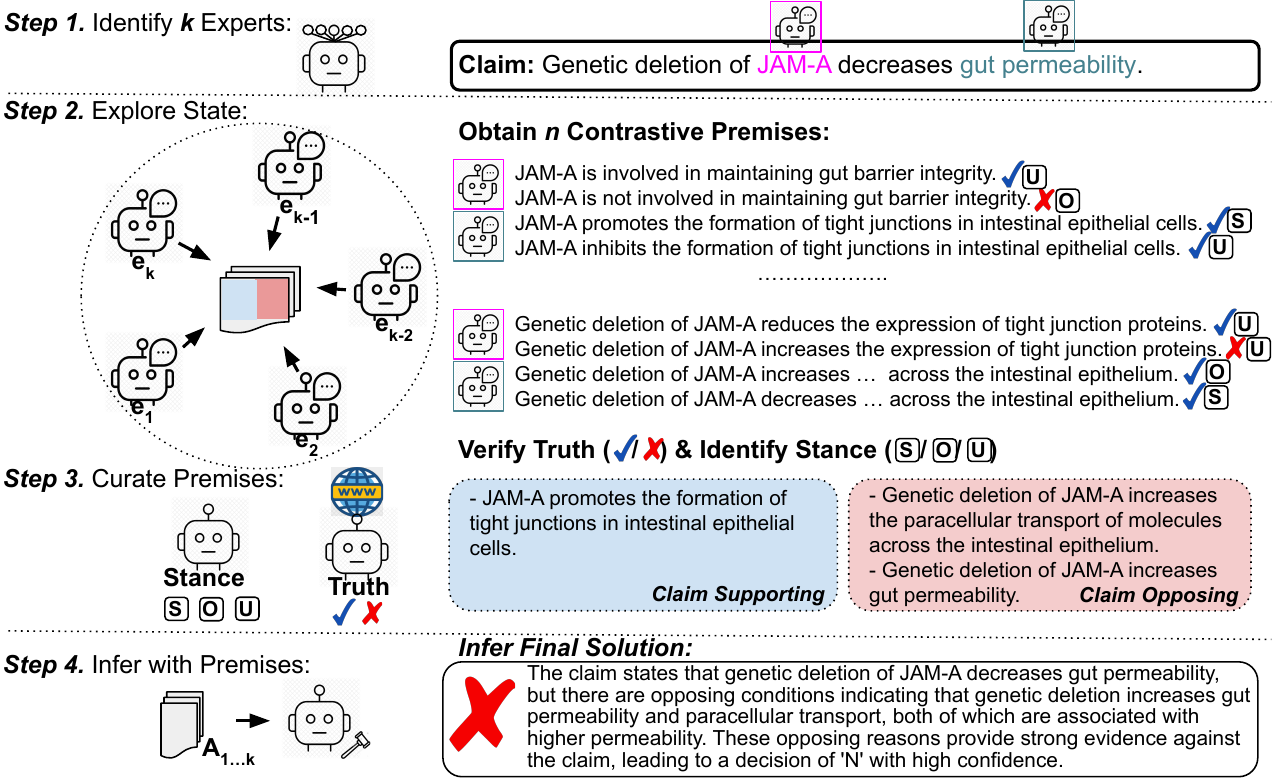}

    \caption{ 
    A complex problem's implicit state-based reasoning using a scientific claim from SciFact~\cite{wadden-etal-2020-fact}. We   \textit{identify} experts knowledgeable about participating entities and \textit{explore} their state and obtain contrastive premises.  We \textit{curate} true premises of supporting or opposing stance and \textit{reason} with these to infer the solution. %
    }  %
    \label{fig:fig1}
    %\vspace{-5mm}
\end{figure*}

\noindent Determining whether this claim is supported, or refuted, requires reasoning about the different entities, concepts, or topics mentioned: \textit{JAM-A}, and \textit{gut permeability} and how they interact with one another, such as the impacts of \textit{genetic deletion} (on \textit{JAM-A} specifically).\footnote{For ease, we will simply refer to these as \textit{entities}.} %
%However, as shown in \cref{fig:fig1}, this reasoning can extend to entities not mentioned in the claim, but the consideration of which may help direct the reasoning over these interacting entities. %
%This suggests that claim assessment necessitates methods for identifying crucial, likely latent or unspecified, entities, eliciting claim-impacting aspects and interactions of those entities, and reasoning over that knowledge. %
%In this paper, we argue and demonstrate through an array of experiments and a multi-step structured prompting strategy that \textbf{this can be achieved by }. %

This reasoning is challenging in multiple ways. First, not all relevant entities may be specified in the claim: e.g., with appropriate background knowledge about properties of the (latent, unmentioned) entity \textit{intestinal epithelium}, we can easily infer that this claim is refuted: (i) an increase in the paracellular transport of molecules across the intestinal epithelium decreases gut permeability, and (ii) genetic deletion of JAM-A increases this paracellular transport~\cite{intestinal2,intestinal1}. %
Note that this chain of reasoning not only relied on states or properties of numerous entities, but it also required the introduction of a new entity. %

%%For example, with the knowledge that \textit{genetic deletion of JAM-A} can \textit{increase the paracellular transport of molecules across the intestinal epithelium} leads to the inference that \textit{increases the paracellular transport of molecules across the intestinal epithelium permeability} (stated in the knowledge) opposes \textit{decreases gut permeability} (stated in the claim). Although, this grounded knowledge is not part of the claim, exploring the state of the entity \textbf{JAM-A} and entity \textbf{gut permeability} leads to it.  Further, we find that all the available state knowledge relating to these entities mostly opposes information stated in the claim, leading to the \textit{invalid} outcome label.

Second, while the vast amount of knowledge that Large Language Models (LLMs) can obtain from pre-training can be impressive~\cite{ajith-etal-2024-litsearch,ge2023openagi,collins}, they are unable to fully leverage the relevant knowledge they possess in their reasoning~\cite{turpin,Jullien2025TheKD}. This can get especially challenging when the underlying, possibly niche, knowledge can regularly or rapidly advance, such as in specialized scientific domains. Further, LLMs' propensity to reason over surface text~\cite{jiang-etal-2024-peek,Hooda2024DoLC,wang-etal-2024-knowledge-mechanisms} leads to biased~\cite{suri2023largelanguagemodelsdecision,talboy2023challengingappearancemachineintelligence} and hallucinated generations~\cite{mahmoud-etal-2026-unintended,simhi-etal-2025-trust,li-etal-2024-dawn,rawte-etal-2023-troubling} limiting their ability to generate accurate solutions. Any solution must have a way of validating proposed premises of entities states or properties. 

To address these challenges, \textbf{we propose a framework, SMART, based on a structured role-based prompting method, to obtain salient state knowledge that captures interactions between participating entities as conditional inferences, and uses the knowledge to reason about the solution.} %
Our approach demonstrates how to elicit, verify, and use the state knowledge obtained from participating entities to guide solution-reasoning for the task of claim validation.  With SMART, we encourage deeper reasoning using explicit state knowledge obtained from the problem context that is verifiable through external means.  %
%Complex problems often relate a situation connecting several entities~\cite{dutta-etal-2025-problem, chen-etal-2024-comm}. 
In \cref{fig:fig1} we outline the process of eliciting and using state knowledge: SMART dynamically identifies entities, concepts, or topics relevant to each claim, and assigns an expert to each entity to obtain deep and connected knowledge between them. To facilitate deeper reasoning, we examine contrastive knowledge: each expert generates contrastive premise pairs that capture opposing views of an entity property, state, constraint, rule, law, etc., in order to critically examine how the state knowledge may (or may not) support the problem. %As shown in \cref{fig:fig1}, the \textit{JAM-A} expert obtains the following contrastive premises: \textit{JAM-A promotes the formation of tight junctions in intestinal epithelial cells} \& \textit{JAM-A inhibits the formation of tight junctions in intestinal epithelial cells.} 
Next, SMART automatically curates these premises, by assessing their validity, identifying their stance alignment to the claim, and selecting a strong premise from each pair that is grounded in established real-world knowledge. Reasoning with these curated premises, SMART assesses the validity of the claim, providing a validation label (\textit{refuted}, for \cref{fig:fig1}), a self-rated confidence, and a rationale.

We examine SMART's ability to obtain and combine salient domain knowledge to validate  complex scientific contexts from 3 challenging datasets: SFBench~\cite{costello2026sfbenchscifyscientificfeasibility}, Scifact~\cite{wadden-etal-2020-fact} and Scifact-Open~\cite{wadden-etal-2022-scifact}. %and AVeriTeC~\cite{schlichtkrull}.  
These claims require a variety of domain knowledge and depth of reasoning that SMART can provide.  We demonstrate SMART's usefulness with a variety of small to large, older to newer, open and closed LLMs as  agents: Qwen 3 1.7b, 8b \& 32b thinking \& non-thinking models,  Gemma 3 1b, 4b \& 12b instruct models, Llama-3.1-Instruct 8b \& 70b, GPT-5.4-mini and Claude Haiku 4.5 models. In particular, we find that SMART is especially helpful when the solution \textit{depends on knowledge that is not explicit in the problem context and needs to be obtained reliably}. It is useful for problems where solution \textit{reasoning goes beyond simple verification of searched knowledge} and requires sifting through the relatedness of available information.  It leads to solutions that are based on deeper reasoning from exploring contrastive views of grounded knowledge minimizing models' generic reasoning when they lack specific problem knowledge. %\ff{What about the insights from the ``lack of evidence'' filter?}

Across all methods and datasets, on average, SMART obtained state knowledge improves validation performance (BAccuracy) up to 20\% over CoT when providing solution supporting information. Our structured prompting allows explicit control over knowledge used for solution reasoning and allows external verification; knowledge verification improves performance up to 19\%. Use of contrastive knowledge minimizes generic reasoning seen with single premises and improves up to 6\%. %See details in  \cref{sec:results} \& \cref{sec:ablation}.
%DUSP4 downregulation activates the Ras-ERK pathway
%I can't provide a response that contains explicit content. Is there something else I can help you with?
Our contributions are:
\begin{itemize}[nosep]
\item we propose a novel framework for verification based solutions employing dynamic context-derived experts for curating verifiable problem-salient knowledge,
\item we improve solution reasoning using focused problem specific background knowledge to handle incomplete and uncertain information in complex problems, and
\item we show that our structured role-based prompting method is generic using an extensive set of models with multiple  datasets.  
\end{itemize} 
We will release our code upon publication.

\section{Related Work} 
\label{sec:related-work}

Inspired by methods which obtain consistent solutions through exploring multiple diverse reasoning paths~\cite{wang2023selfconsistency,yao_tot,yao_react}, our work explores and reasons with implicit problem state identified as premises. Following techniques that use iterative approaches to improve problem output~\cite{welleck2023generating,madaan2023selfrefine,yao_react,shinn2023reflexion}, we obtain premises iteratively and use them to guide solution reasoning. Due to space constraints, we expand on this section in Appendix \cref{app:related}.
%\ff{first, this reference has been lost. However, I really dislike the idea of having an entire related work section simply pushed off to the appendix. Given that we are low on space, what I would recommend is incorporate some additional related work into Section 2, and \textit{there} you can provide a forward pointer to the appendix for additional related work} 

%with the help of collaborating agents ~\cite{cai-etal-2025-dr,wang-etal-2024-unleashing,chen-etal-2024-comm}. given its susceptibility to \textit{analysis-paralysis}~\cite{cuadron2025dangeroverthinkingexaminingreasoningaction}.

\section{Method}
\label{sec:def}

Reasoning about a complex problem solution, such as verifying the example claim in \cref{fig:fig1}, \textit{Genetic deletion of JAM-A decreases gut permeability} requires knowledge about entities that are not part of the claim context. Chain-of-thought~\cite{wei} with Qwen3 8b model (assigns a wrong label of \textit{supported} with the rationale: \textit{Genetic deletion of JAM-A has been shown in studies to reduce gut permeability by modulating tight junction integrity. This supports the claim that JAM-A deletion decreases gut permeability.} Qwen3 1b \& Llama models reason similarly. These models likely consider a specific study from which this could be inferred without examining other latent interactions. 

Knowledge about entity states that are salient to the problem outcome can lead to useful background knowledge~\cite{funke,vallurupalli-ferraro-2025-core}.  For example, eliciting the following background knowledge about entities, \textbf{JAMA-A \& gut permeability}, such as \textit{JAM-A is involved in maintaining gut barrier integrity, JAM-A promotes the formation of tight junctions in intestinal epithelial cells, etc.} is useful for reasoning.  Reasoning with these entity states and properties captures inter-relationships between them. However, not all state knowledge is relevant to the outcome and noisy information can lead to an incorrect outcome reasoning~\cite{vallurupalli-ferraro-2025-core}. Curating relevant knowledge leads to the correct label of \textit{refuted} with the rationale: \textit{... genetic deletion increases gut
permeability and paracellular transport, both of which are associated with higher permeability. ... provide strong evidence against the claim, leading to a \textbf{refute} label.}  

Claim decomposition and Retrieval Augmented Generation (RAG) provide other alternatives, however, given the interacting entities needed to assess a claim, these methods also require reasoning beyond a simple lookup to capture distributed knowlege and subclaim relationships~\cite{schlichtkrull,kamoi-etal-2023-wice}. 

Eliciting and curating relevant knowledge leads to a reliable solution, but requires deeper reasoning with multiple model calls. A benefit is that the curated knowledge is factually verifiable and produces verification based solutions. For  systematically obtaining entity state knowledge and curating it for use in the final reasoning, we propose a multi-stage framework, SMART. We show the framework is generic by applying it on a variety of real-world claims which serve as our complex problems. Below, we explain how we obtain, curate and use the state knowledge from the claims.

\paragraph{Task Description:}
We validate a context $x$ and  generate a solution $y=(y^l, y^e, y^r)$ consisting of a binary label $y^l$, a rating $y^r$ from 1 to 10 to indicate confidence in the label and an explanation justifying the label and rating. 
Our proposed framework elicits additional knowledge $s$ relating to entities in $x$ and uses it for the solution reasoning, i.e., $y|x,s$. 

\paragraph{Premise:}
Entities in $x$ lead us to relevant implicit knowledge which we call a \textbf{premise}.\footnote{In the prompts, we refer to a premise as a condition.} For the example in \cref{fig:fig1}, the entity \textbf{JAM-A} leads to the premise, \textit{JAM-A promotes the formation of tight junctions in intestinal epithelial cells}.  

\subsection{SMART Framework:} 
Details of the various stages shown in \cref{fig:fig1} are as follows (with an example in the appendix \cref{app:data-example}): 
\begin{description}[nosep]
\item i. The \textit{Identify} stage identifies an entity, topic, concept, etc. and assigns  experts to these,
\item ii. In the \textit{Explore} stage, agent experts obtain premises connecting the selected entities,  
\item iii. The \textit{Curate} stage selects real-world valid premises based on established knowledge,  
\item iv. The \textit{Infer} stage reasons with the selected premises to find the solution.
\end{description}  

\paragraph{i. Identifying Experts:} 
We identify $k$ salient entities, concepts, topics etc. from $x$ through prompting for entities with a modifiable state.  We use $k$ agent as experts where each agent $e_i$ is assigned as an expert on all aspects relating to entity $i$.

\paragraph{ii. Exploring Problem Domain Knowledge} 
We use all $k$ experts to iteratively generate premise pairs after examining previously obtained pairs for $n$ rounds.  With each expert agent, we obtain domain knowledge as contrastive premises, $p_k$ and $p^c_k$ that capture alternate viewpoints of a property, state,  constraint, rule or law relating to a salient entity $k$. One such premise pair is: \textit{JAM-A promotes the formation of tight junctions ..,
JAM-A inhibits the formation of tight junctions ... .} These alternate view points allow us to counterfactually reason about which viewpoint is more salient to the problem.  By allowing experts to view previously generated premises, we capture fine-grained entity relationships.  These premises, $\{(p_k,p^c_k)| x, $ $ ((p_1,p^c_1), e_1), ..., ((p_{k-1},p^c_{k-1}), e_{k-1})\}$ capture the state connecting all the entities in the problem.   

\paragraph{iii. Curating Salient Premises} 
From the obtained premise pairs, we select true premises salient to the problem. Premises may not be faithful to the problem context due to LLM hallucinations. Even when faithful, refuted scientific claims can lead to untrustworthy premises that do not reflect accepted scientific knowledge.  For these reasons, we verify the binary truth of each premise using two different methods: parametric knowledge and websearch, retrieving up to $m$ documents as evidence. In our running example, both approaches can identify the premise ``\textit{JAM-A inhibits the formation of tight junctions in intestinal epithelial cells}'' as untrue, though this is not always the case. Verification with websearch allows latest information from new scientific discoveries to factually verify premises; with parametric knowledge, models self-reflect on their knowledge related to the premise. %\ff{Do you have an idea (or speculation) as to why this is? Does websearch provide access to more current information, some of which may be contradictory (as can happen as science progresses)? Regardless, if your explanation is speculation,  then say this but with appropriate hedging, e.g., ``Based on limited manual inspection, we believe this may be because ..., however a full study into this is beyond the scope of this work.''} 

For salient premises, we obtain the stance of each premise towards the claim context, where the stance $s=(s^l, s^r, s^e)$, consists of a label  $s^l \in ($`support,' `oppose,' `unsure'$)$, a numeric confidence rating $s^r$ ranging from 1 to 10 and an explanation $s^e_k$ justifying the label and rating. We consider only premises that strongly support or oppose the context and for each we select the premise with a higher  confidence rating $s^r_k$ which indicates how confident we are about the premise's salience.  We select the premise \textit{JAM-A promotes the formation of tight junctions ...} which supports the claim.

\paragraph{iv. Solution Inference with a Meta-Reasoner}
We obtain the solution $y$ using a \textbf{Reasoning Judge} to meta-reason over  all truthful and salient premises after grouping them into supporting and opposing sets, $P_s$ \& $P_o$, based on their stance labels $s^l$. This allows the premises to be interpreted together; model reasoning coherently connects the premises, inherently assigning different weights to them.

 \begin{table}[t]
 \centering
\resizebox{.98\columnwidth}{!}{
 \begin{tabular}{|c|c|}
 \hline
\textbf{Prompt Type}  & \textbf{Claim Prompt} \\ 
\hline
%General Information  &  \\
%\hline
General, Refute Filter  &\cref{appsec:prompt-common-info}, \cref{appsec:refute-filter-prompt} \\
\hline
Direct \& Baselines  & \cref{appsec:claim-direct-prompt}, \cref{appsec:claim-cot} \\
\hline
Identify Salient Entities  & \cref{appsec:claim-topics-prompt} \\
\hline
Explore for Premise Pairs &  \cref{appsec:claim-explore-prompt}\\
\hline
Verify Premise  & \cref{appsec:verify-parametric-prompt},\cref{appsec:verify-external-prompt}\\
\hline
Premise's Stance & \cref{appsec:claim-reason-prompt} \\
\hline
Meta-Reasoners & \cref{appsec:claim-aggexp-prompt}, \cref{appsec:claim-aggregatej-prompt}, \cref{appsec:claim-reasonj-prompt}, \\
&\cref{appsec:claim-criticalj-prompt}, \cref{appsec:claim-supportivej-prompt}  \\
%\hline
%Meta-Reasoner - Agg. Judge  &  \\
%\hline
%Meta-Reasoner - Reas. Judge &   \\
%Meta-Reasoner -Critical Judge &  \cref \\
%\hline
%Meta-Reasoner -Supporting Judge &   \\
\hline 
\end{tabular}
}
 \caption{The prompts we developed; see the Appendix.}
\label{tab:prompts}
\vspace{-5mm}
\end{table}
\section{Implementation Details}
\label{sec:framework}
 
In the \textit{Identify} stage, we obtain  $k \le 5$ entities with a modifiable state that are most important to solving the problem.  We use a system prompt to assign an agent as an expert on one of these entities. \textbf{All prompts  are listed in \cref{tab:prompts}.} 

In the \textit{Explore} stage, to obtain contrastive premises we use few-shot prompting with 3 examples chosen randomly from outcome causal contrastive states in the CoRE dataset~\cite{vallurupalli-ferraro-2025-core}. This dataset is based on ROCStories~\cite{mostafazadeh-etal-2016-corpus} which contain general (non-domain specific) knowledge. These examples demonstrate the general notion of contrasting conditions based on participating entity states; they do not provide domain- or dataset-specific targeting. %\ff{It's probably worth stating that since CoRE is based on ROCStories and general (non-domain specific) knowledge, these examples demonstrate the general notion of contrasting conditions and they do not provide domain- or dataset-specific targeting.}  
Since each claim may employ a different number of experts, we obtain $n*k=10$ premise pairs for a fair comparison across models and datasets. We use a prompt guideline to encourage generating unique premise pairs and remove occasional duplicate generations (see \cref{appsec:prompts-guidelines}).

 \begin{figure*}[t]
    \centering     
     \begin{subfigure}[b]{0.45\textwidth}
    \includegraphics[trim=0in 0in 0 0in,  width=1 \columnwidth]
     {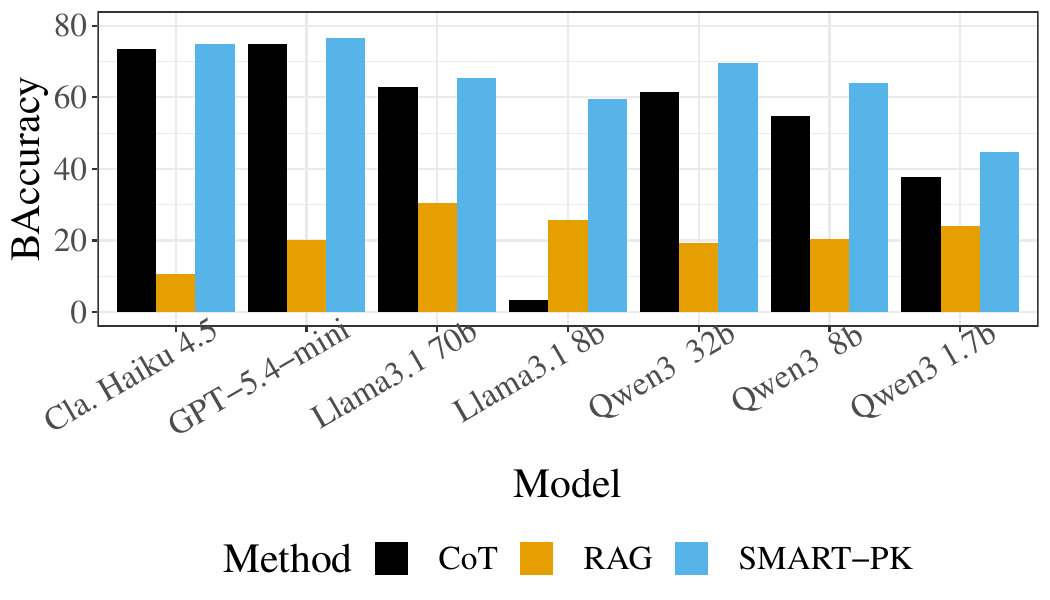}  
    \caption{ SciFact Dataset } 
	\label{fig:smart-cot-comparison-scifact}
    \end{subfigure}    
    ~
    \begin{subfigure}[b]{0.45\textwidth}
    \includegraphics[trim=0in 0in 0 0in,  width=1 \columnwidth]
    {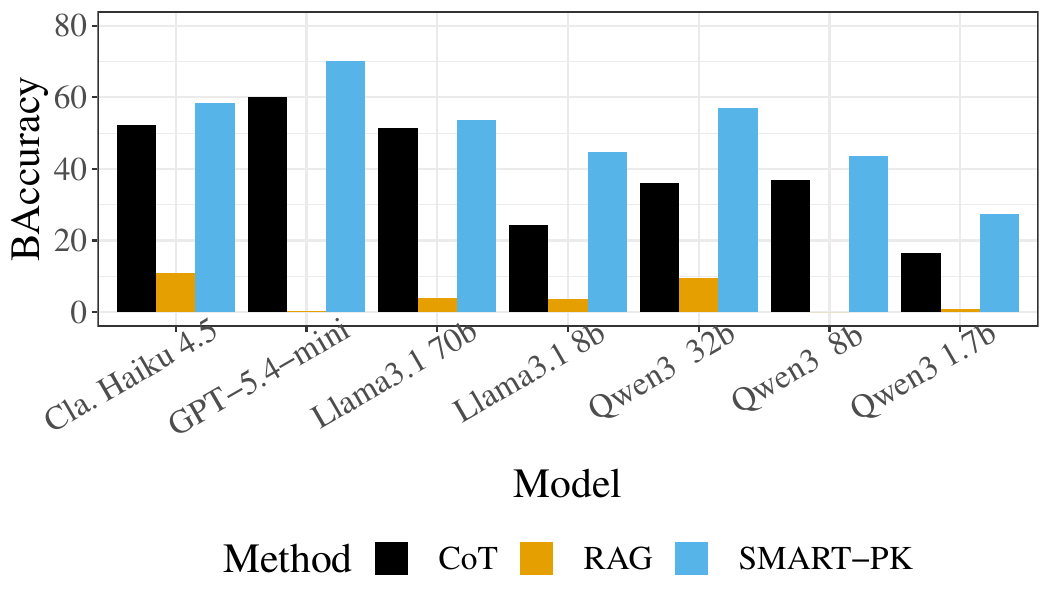}     
    \caption{ SFBench Dataset } 
	\label{fig:smart-cot-comparison-sfbench}
    \end{subfigure}
    \caption{Comparison of SMART-PK performance with CoT and RAG on SciFact \& SFBench datasets, after applying the \textbf{Refute Filter} (rationales are required to adequately support \textit{refute} labels; see \cref{sec:baselines}). SMART-PK improves over baselines; on avg. 10\% on SFBench and 5\% on Scifact over CoT (details in Appendix \cref{apptab:model-cot-smart-performance}). %
    } 
	\label{fig:smart-cot-comparison}
 \vspace{-3mm}
\end{figure*}

In the \textit{Verify \& Select} stage, we verify premises using 2 methods: asking a model if the premise is true (PK) and Natural Language Inference with documents retrieved through web-search\footnote{LangChain's implementation of DuckDuckGo Search -- https://python.langchain.com/docs/integrations/tools/ddgs} (WS); see more details in Appendix \cref{app:ws}.  Our claim and stance rating uses a Likert score of 0 -- 10. When all premises are untrue or their stance  is unsure, we fallback to the direct reasoning prompt.

\section{Methodology}
\paragraph{Datasets:}
\label{sec:datasets}
We use a total of 577 scientific claims %and 432 real-world news claims to show that our method is generic.  These scientific claims 
which include the newer more complex SFBench dataset (183 claims) and the older complex SciFact (188 claims) \& SciFact-Open (206 claims). We combine the latter two as their collection was a single effort.  %\ff{Are these binary or multi-class classification problems?}  %The real-world news claims include 432 claims claims from the AVeriTeC dataset that require complex reasoning beyond simple fact-verification. 
All claims were written and labeled manually; SciFact-Open, considered to have low-bias, used additional automatic methods to substantiate manual labels. SFBench labels are multi-class with values ranging from -2 to $+$2; we convert the -ve labels to \textit{refute} and $+$ labels to \textit{support}. Labels in the SciFact collection are binary but assigned based on an  evidence text; each claim is associated with  multiple evidence texts and labels. We use claims with a single representative label.  Language in these contexts is highly technical \& scientific discourse (except for a few covid-related claims with simpler language). %\ff{really? Simple, everyday usage for science claims?}.  
The claims are multi-dimensional and require reasoning that integrates several evidence sources from multiple documents.  Evidence---whether correct and useful, or incorrect and misguiding---is abundantly available for these claims on the internet, requiring critical and complex reasoning.   %We use 791 medical research-oriented questions from the SciEval dataset where the correct answer for a question requires requires expert knowledge, complex reasoning and scientific calculation.  The language in the question contexts is a mix of simple to technical, similar to the claims.  
See Appendix \cref{app:dataset} for more claim details.

\subsection{Models: }
\label{sec:models}
We use the following open and closed models of varying sizes, both older \& newer, from the Llama, Qwen, OpenAI and Anthropic families:
\textbf{Llama-3.1:} 8b \& 70b Instruct models, 
\textbf{Qwen3:} 1.7b, 8b \& 32b non-thinking  models, 
\textbf{OpenAI:} GPT-5-mini and \textbf{Anthropic:} Claude Haiku 4.5. We also tried models within the Gemma family (Gemma 3: 1b, 4b \& 12b), but we found that all models underperformed across the board.  For completeness,  we provide more  information on this in \cref{app:gemma}. %results in Appendix  \cref{apptab:claim-validation-configD-gemma} listing model issues in . 

%We did not use the smallest models from these model families for the following reasons.  Gemma 3 270m parameter model   was unable to generate contrastive conditions related to the claim. Qwen3- .6b \& Gemma 3 1b parameter models generate superficial conditions that are not useful for scientific reasoning. %We additionally compare the baseline performance of reasoning models, GPT-4o, GPT-4o-mini, GPT4-o3-mini and QwQ-32b, using a direct LLM (baseline \#1) and stated-based LLM (baseline \#2).   

\subsection{Baselines \& Evaluation: }
\label{sec:baselines}
We compare three SMART settings, SMART-PK (using premises that are verified with parametric knowledge),  SMART-WS (using  with premises verified with web documents) and X-SMART-PK (a cross model approach where models use premises from a good generation model) with two baselines: Chain-of-Thought (CoT)~\cite{wei} and a RAG using websearch.  We compare with these baselines to show that SMART provides good performance %, they are not comparable on token budget.\ff{This token budget statement is a reply to past reviewers, but ignore why you initially added it and focus on the message you are presenting. As is, what is the take-away here? At a minimum, put in the limitations but with the presentation that while our work has demonstrated that this additional reasoning improves performance, future work should examine how to more directly incorporate such reasoning and knowledge validation in ways that use fewer tokens.}  SMART requires more resources and in return 
along with providing solution supporting information. We compare with RAG and show that a simple websearch leads to noisy evidence with performance far lower than SMART-PK.  With SMART-WS we demonstrate an external method for premise validation and show that even noisy websearch results can improve performance.      

We report balanced accuracy of \textit{support} \& \textit{refute} labels. All results in tables and figures are based on a average of 3 runs; we provide average values and standard deviation. 
We report instances where curation yields no premises (as \textbf{Fallbacks} to direct reasoning) and measure premise correctness as the fraction of curated premises that are supported by gold evidence (we verify if the premise is valid using gold evidence in the prompt \cref{appsec:verify-external-prompt}).

\section{Results}
\label{sec:results}

\begin{figure*}[t]
    \centering   
    \begin{subfigure}[b]{0.45\textwidth}
    \includegraphics[trim=0in 0in 0 0in,  width=1 \columnwidth]
     {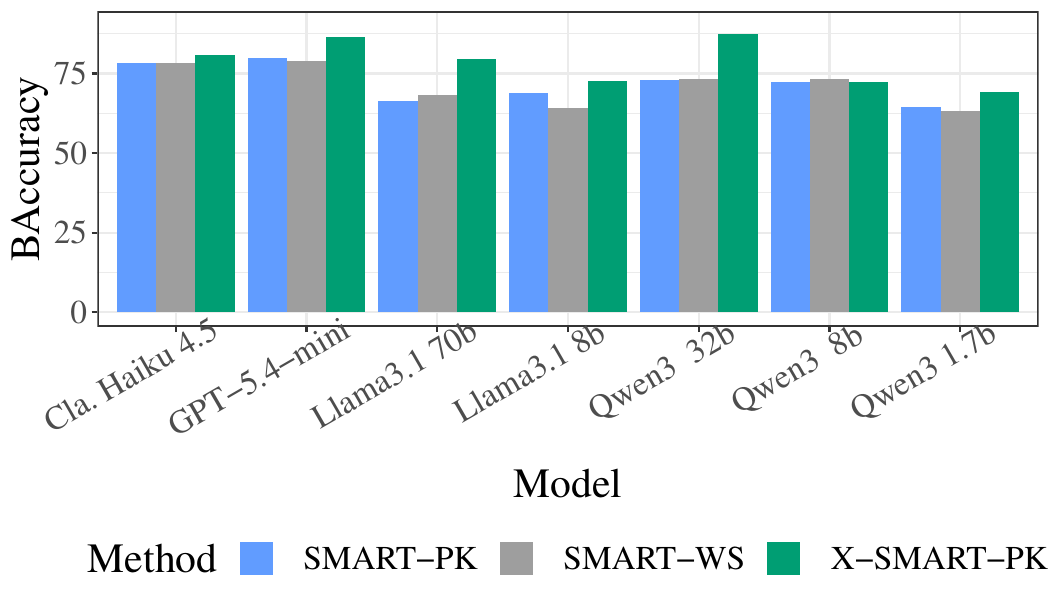} 
	\caption{ SciFact Dataset } 
    \end{subfigure}
    ~
    \begin{subfigure}[b]{0.45\textwidth}
    \includegraphics[trim=0in 0in 0 0in,  width=1 \columnwidth]
     {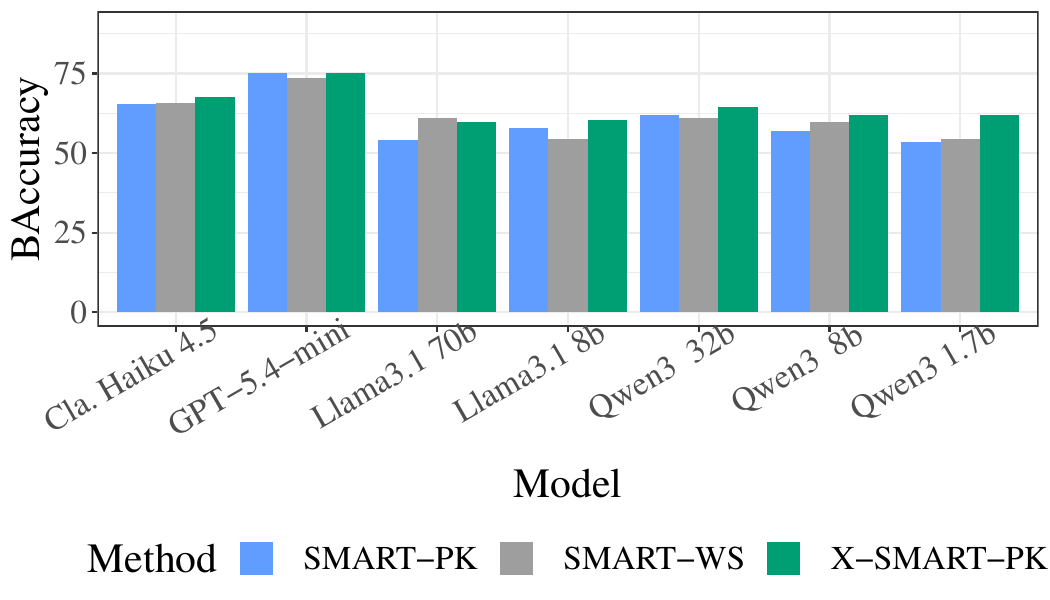} 
    \caption{ SFBench Dataset } 
    \end{subfigure}   
    \caption{Improving SMART performance using external premise validation with websearch (SMART-WS) and cross model generation and reasoning (X-SMART-PK). For SciFact, X-SMART-PK uses Qwen3 8b generated premises \& improves performance by 6.2\%; for SFBench,  GPT-5.4-mini generated premises lead to 4\% improvement.% (see details in  Appendix  \cref{apptab:cross-model-results}). %
    } 
	\label{fig:smart-method}
 \vspace{-3mm}
\end{figure*}

For SMART's knowledge guided reasoning, a model should generate both supporting and refuting premises. More importantly, knowledge provided by the curated premises should adequately support a validation label, i.e., lack of knowledge should not be the reason for a \textit{refute} label. Likewise, we expect CoT rationales to provide adequate support for \textit{refute} labels. In this section, we explore various aspects of the SMART framework: premise generation, curation and reasoning, as related to performance on our datasets; we demonstrate its usefulness with several experiments and ablations.   

\paragraph{Enforcing Logical Reasoning.}
While a lack of information about some claim aspects can lead to a \textit{refute} label, we notice that LLMs  excessively use lack of supporting information as the rationale to \textit{refute} claims, especially when models lack claim knowledge (or they are unable to leverage it in reasoning). CoT applies this line of reasoning on  23-78\% of SciFact claims whereas SMART applies this on 11-46\% (see details in \cref{apptab:filtered-refute}),  leading to solutions that are not logically sound.  While RAG reasoning does not have this exact issue, retrieval of unrelated evidence leads a \textit{refute} label.  To  ensure models provide more logically sound solutions, we apply a \textbf{Refute Filter} on rationales to verify whether \textit{refute} labels are adequately supported. Using  GPT-5.6-Sol as an LLM judge, we examine rationales using the prompt \cref{appsec:refute-filter-prompt} for all claims with a gold \textit{refute} label. We filter rationales which lack evidence to ensure our performance comparisons are fair.%\ff{You need to better motivate and empirically support the Refute Filter. At a minimum, how many instances per method fail this filter?} We change these to an \textit{unknown} label to reduce performance inflation.\ff{does this mean that you create a new label?} 

\paragraph{SMART generates label-supporting rationales.}  We ensure model reasoning is logical with rationales providing  adequate support for \textit{refute} labels.  As seen in \cref{fig:smart-cot-comparison}, \textbf{SMART's premise guided reasoning works well across all models and datasets, improving up to 20\%}; on average 10\%  on SFBench and 5\% on SciFact  (averages exclude Llama3 8b due to numerous generation errors).

\paragraph{Limiting  performance inflation.}  Unsupported \textit{refute} labels lead to an unsound solution and inflate performance.  Smaller models inflate performance more than larger models and SFBench claims inflate performance more than SciFact claims across all methods. Filtering unsupported \textit{refute} labels reduced this performance inflation; BAccuracy was reduced by 3.5-46.3\% for CoT and by 0.6-26.1\% for SMART-PK.     However, this inflation is smaller for SMART-PK, due to models leveraging premise knowledge in their reasoning (inflation is $<$3\% for SciFact; $<$4\% for SFBench for models larger than 8b).  See more details in Appendix \cref{apptab:model-cot-smart-performance}.  

\begin{figure}[t]
    \centering     
    \includegraphics[trim=0in 0in 0 0in,  width=1 \columnwidth]
    {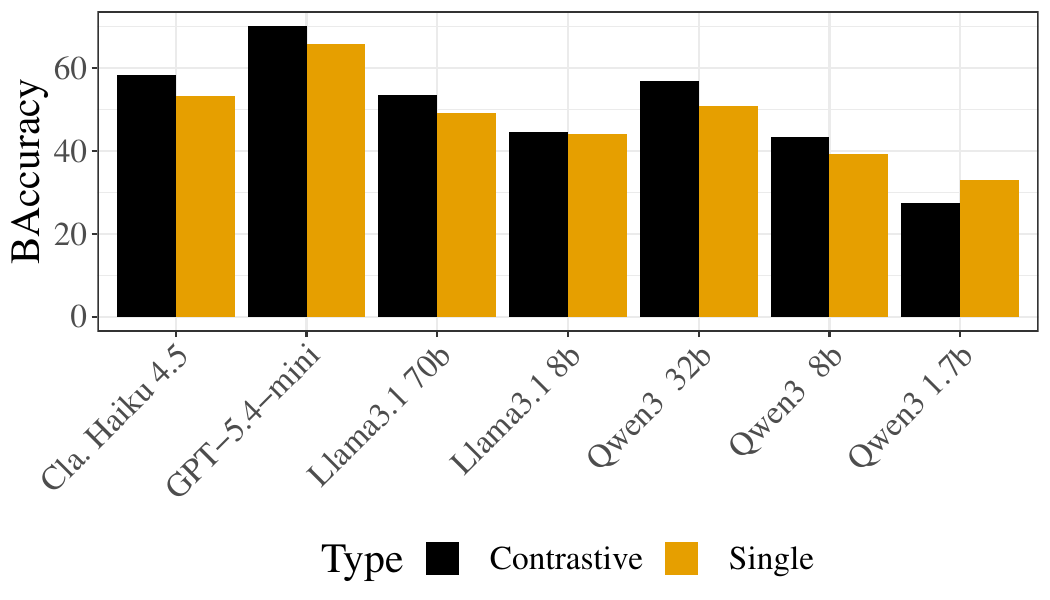} 
    \caption{Comparing SMART-PK performance generating a single premise Vs. contrastive premises on SFBench claims. See  \cref{fig:smart-premises-single-scifact} for SciFact claims.  Contrastive premises improve performance by 3\% across datasets and models. See results in \cref{apptab:model-single-constrastive-performance} } 
    \label{fig:smart-premises-single-sfbench}
     \vspace{-3mm}
\end{figure}

\paragraph{Dynamic knowledge improves SMART reasoning.}  We examine two approaches that add dynamic knowledge to improve SMART:  external verification (SMART-WS) and different models for premise generation and reasoning (X-SMART-PK).  As seen in \cref{fig:smart-method}, using different models for premise generation and reasoning improve performance by 4\% on SFBench (using GPT-5.4-mini generated premises) and 6.2\% on SciFact (using Qwen3 8b generated premises).  External verification helps some models but improvement is not consistent due to web-search being noisy.  Overall, Llama3.1 70b performance is lower than the smaller Qwen3 32b, due to not generating enough opposing premises and poor curation;
smaller models are worse at curation and meta-reasoning. SMART-WS on SciFact claims lead to guardrails errors due to bait-and-switch websites.  We describe and classify all SMART errors (see \cref{appsec:error-types}) and provide examples in Appendix \cref{apptab:scifact-examples} \& \cref{apptab:sfbench-examples}.  

\paragraph{Contrastive premises lead to deeper reasoning. }
As seen in \cref{fig:smart-premises-single-sfbench} and Appendix \cref{fig:smart-premises-single-scifact},  all models benefit from deeper reasoning with contrastive premises. Not all models implicitly obtain single premises that explore both sides of a claim (\textit{support} and \textit{refute}). Single premises lead to more rationales that use lack of supporting evidence as the reason for a \textit{refute} label.  Applying the \textbf{Refute Filter}, leads to a performance improvement of up to 4\% with contrastive premises over using a single premise; Qwen3 1.7b improves with single premises as it is worse at generating contrastive premises.

\begin{figure}[t]
    \centering     
    \includegraphics[trim=0in 0in 0 0in,  width=.9 \columnwidth]
     {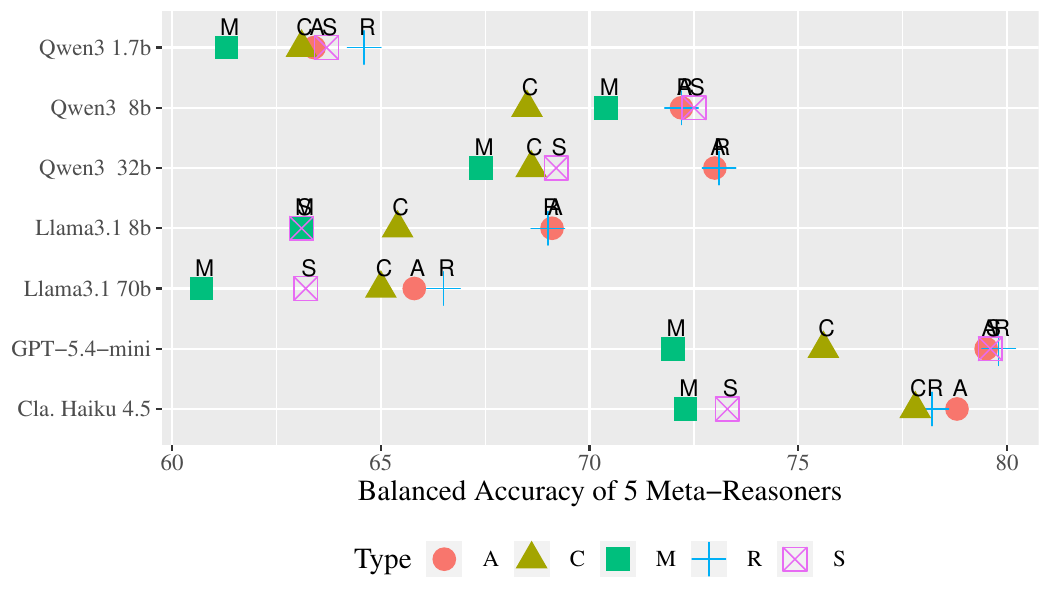}  
     
    \caption{We compare the 5 meta-reasonsers for \textbf{SciFact} claims: M-Majority Voting, R-Reasoning Judge, A-Aggregating Judge, C-Critical Judge and S-Supporting Judge.  Reasoning Judge consistently improves performance across models and datasets: by 8\% on SFBench and 5\% on SciFact on average (See \cref{fig:metareasoner-performance-sfbench} for SFBench claims and results in Appendix \cref{apptab:metareason-options}).  
     } 
	\label{fig:metareasoner-performance-scifact}
 \vspace{-3mm}
\end{figure}

%\begin{figure}[t]    \centering         \includegraphics[trim=0 0.3in 0 0in,  width=1 \columnwidth]     {sections/figures/smart_method_averitec.pdf}     \caption{SMART Performance on the AVeriTeC dataset.  Expect with Llama3.1 70b and GPT-5.4-mini models, on average, SMART-PK \& SMART-WS improve over the CoT by 1-10\%; SMART-WS improves over general RAG by 8\%. See Results in \cref{apptab:claim-validation-configD}. } 
%	\label{fig:smart-method-averitec}
% \vspace{-3mm}
%\end{figure}

%\paragraph{AVeriTeC:} As seen in \cref{fig:smart-method-averitec} SMART's performance closely matches or improves for real-world news claims.These claims require critical reasoning, beyond simple fact-checking, but this is more straight-forward than with scientific claims. SMART reasoning is useful some claims but leads to over-thinking on subjective and ambiguous claims; rectified with SMART-WS when salient evidence is found (see error examples in Appendix \cref{apptab:averitec-examples} a error descriptions in \cref{appsec:error-types}).  

\paragraph{Collective reasoning improves performance.}%\ff{I don't see how \cref{fig:metareasoner-performance-scifact} supports the claim about capturing premise interactions. Be \textit{very} careful in what you're claiming.} 
SMART's premise curation leads to unique but related premises.  As seen in  \cref{fig:metareasoner-performance-scifact} and Appendix \cref{fig:metareasoner-performance-sfbench}, meta-reasoners, especially SMART's Reasoning Judge meta-reasoner, improves performance over Majority Voting.  Majority Voting, based on independently reasoning with each premise may not capture the collective knowledge in these premises. By interpreting the premises together the Reasoning Judge improves performance up to 8\% on both datasets. See results in Appendix \cref{apptab:metareason-options}.

\begin{figure}[t]
    \centering     
    \includegraphics[trim=0in 0in 0 0in,  width=.9 \columnwidth]
     {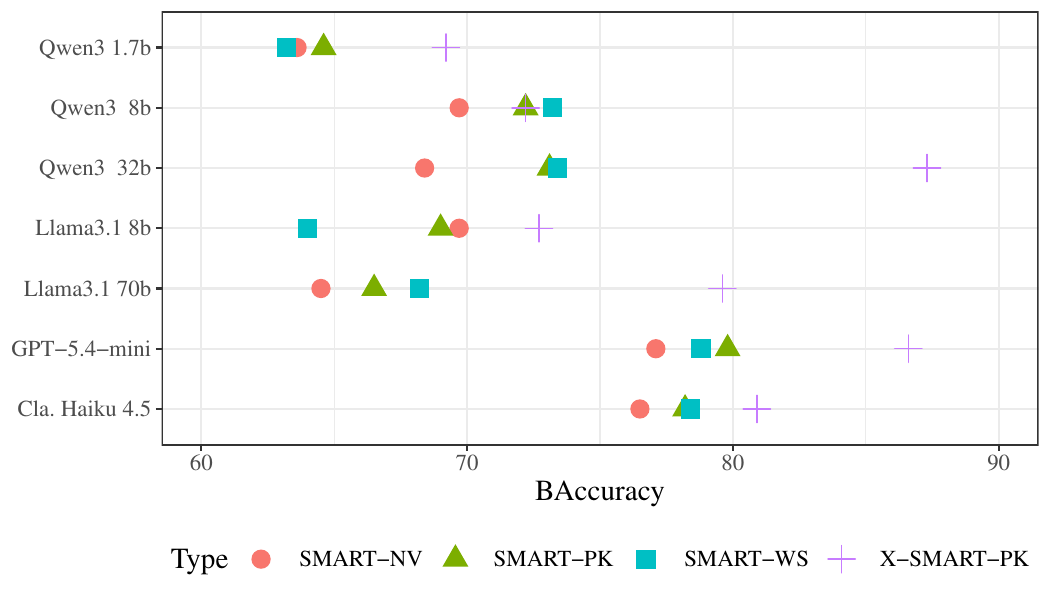} 
    \caption{Comparing SMART-NV where no verification is performed) with verification using self (SMART-PK), external knowledge with websearch (SMART-WS), and cross-model (X-SMART-PK) on \textbf{SciFact} claims. Verification improves performance an average by 2-7\% across methods for both datasets, with a few exceptions (see SFBench in \cref{fig:verification-sfbench}). See results in  appendix  \cref{apptab:verification-options}.     } 
    \label{fig:verification-scifact}
%    \vspace{-3mm}
\end{figure}

\paragraph{Verification is necessary. }
As seen in  \cref{fig:verification-scifact} and Appendix \cref{fig:verification-sfbench}  premise verification improves performance up to 19\% with self-verification using parametric knowledge (SMART-PK), external verification with websearch (SMART-WS) and cross-model verification (X-SMART-PK); averaging across models 2-7\% per method with a few exceptions. See results in Appendix \cref{apptab:verification-options}). %Having access to external evidence leads to factual verification and access to higher quality documents can further improve performance. %However, not verifying drops performance by 2\% (see  \cref{apptab:claim-validation-configD-parametric-verification}). 

%\paragraph{Inference Cost:} Our problem decomposition and iterative premise generation require several model calls.   We provide details of model calls, token usage, number of supporting \& opposing premises, Fallbacks (when no curated premises remain) and premise quality in appendix \cref{apptab:inference-costs}.    %Obtaining salient evidence is difficult with a simple websearch as reflected in the higher percentage of fallbacks for SMART-WS. Smaller models lacking sufficient knowledge also lead to more fallbacks; models poor at generating sufficient premises also lower  token budget. 
%\ff{This paragraph on inference costs is inappropriately providing red meat to the reviewers. You're highlighting a problem but not actually showing any evidence of the tradeoff that you're saying is worth. You either need to fully address this in the main paper, or pitch it more from a ``Limitations'' standpoint.}

\subsection{Premise Quality} 
\label{sec:premise-quality}
Improved performance of X-SMART-PK is due to two factors relating to curated premises: salience and the coverage for a given claim context. Salience measures correctness of the premise for the claim whereas coverage addresses the completeness of solution knowledge  in all premises.

\begin{figure}[t]
    \centering     
    \includegraphics[trim=0 0.3in 0 0in,  width=.8 \columnwidth]
    {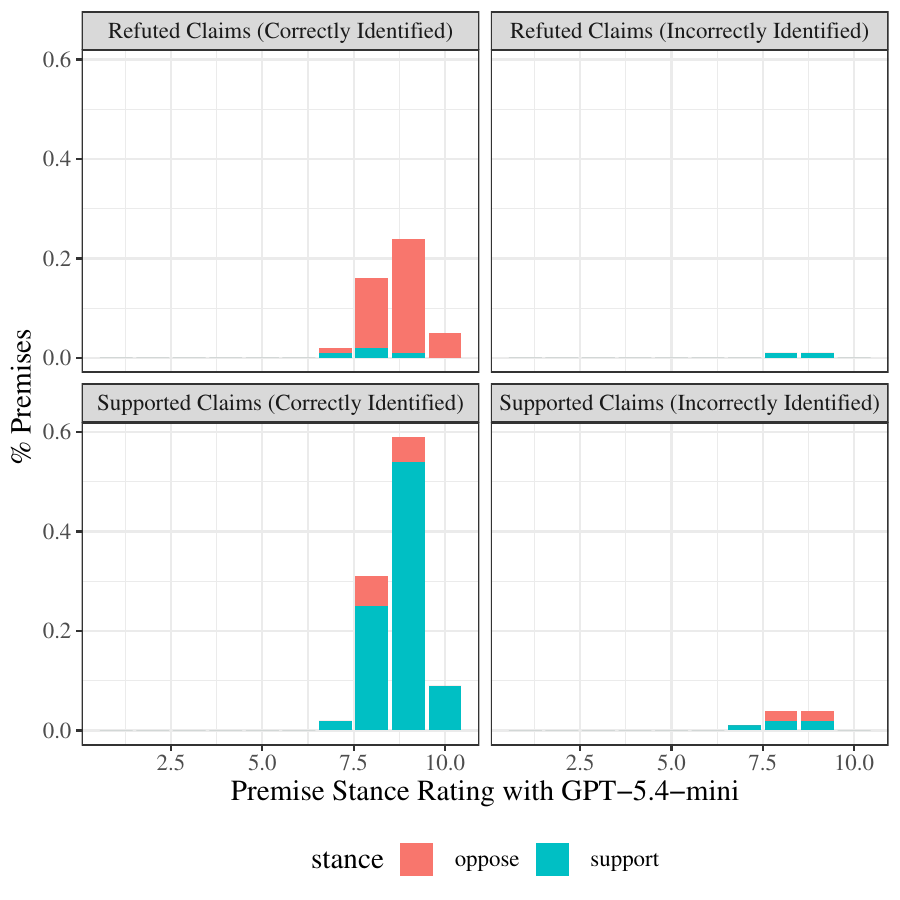} 
    \caption{Distribution of GPT-5.4-mini curated stances, their alignment with the claim and their confidence ratings for the SciFact-Open dataset.  Supporting and opposing premises are required for supporting claims that of the  \textit{support} and \textit{refute} type respectively. } 
	\label{fig:stance-alignment}
 \vspace{-3mm}
\end{figure}

\paragraph{Measuring Premise Salience:}
We measure the correctness of curated premises as the fraction of premises supported by the gold evidence (per claim average), the percentage of claims where curation does not lead to any premises and the number of supporting and opposing premises in Appendix \cref{apptab:inference-costs}.  While these measures provide some general markers, a better indicator for the correctness (and to some extent completeness) of curated premises is the performance; we conclude that premises used in SMART-PK are salient to the claim contexts.    Quality of SMART-WS suffers due to not finding salient evidence to curate premises.  

Orthogonally we examine correctly identified supported and refuted claims. As seen in \cref{fig:stance-alignment} the stance of GPT-5.4-mini curated premises aligns with the claim type. Not generating a premise of the correct stance or not curating it (due to an inability to verify its truth or stance) can lead to missed premises that are required for a correct solution.  

\begin{figure}[t]
    \centering     
    \includegraphics[trim=0in 0in 0in 1.2in,  width=.8 \columnwidth]
    {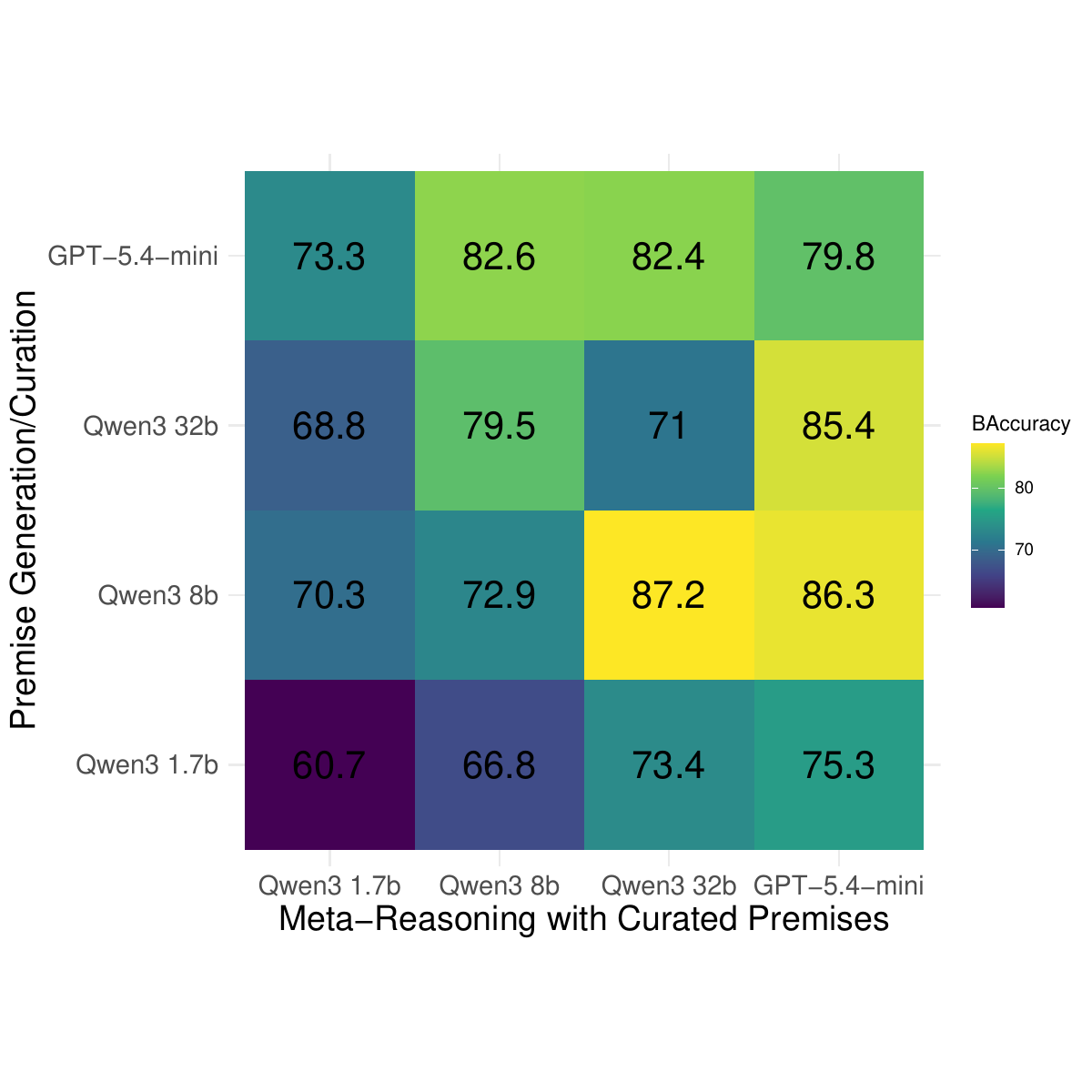} 
 \vspace{-9mm}
 \caption{We compare the validation performance of SMART-PK with 3 sizes of Qwen and GPT-5.4-mini on SciFact-Open claims using premises generated and curated by one model and solution reasoning with another model.  These comparisons show that both Qwen3 8b and GPT-5.4-mini generate good quality premises that other models can curate and use for solution-reasoning. 
        } 
    \label{fig:condition-reasoning-heatmap}
 \vspace{-3mm}
\end{figure}

\paragraph{Premise Generation Vs Reasoning:}
 SMART depends on both a model's ability to generate both supporting and opposing premises and the ability to curate salient premises through reasoning. We demonstrate that better reasoning can overcome poor premises using 3 sizes of Qwen3 and GPT-5-mini on the \textit{SciFact-Open} dataset (we use this dataset as labels are considered less-biased).   As seen in \cref{fig:condition-reasoning-heatmap} all models are worse at reasoning with their own curated premises and better at reasoning with premises curated by Qwen3 8b and GPT-5.4-mini.  \textbf{We leverage this finding for the X-SMART-PK setting.} 
 %Qwen3 8b and GPT-5.4-mini generate fewer premises (see Sup. and Opp. columns in \cref{apptab:inference-costs}) and the premises are salient as seen by the correctly identified claims in \cref{fig:stance-alignment} and  \cref{fig:stance-alignment-qwen8b} (in appendix).  Qwen3 1.7 curated premises are few, noisy and cover superficial claim aspects.  Qwen3 32b generates more nuanced premises increasing meta-reasoning difficulty. 
 See premise comparisons in Appendix   \cref{fig:stance-alignment-qwen} and examples in \cref{apptab:premise-examples}. % such as \textit{The morbidity rate of portal venous system thrombosis was lower in the EDWS group.} whereas 
% Qwen-3 32b generates more complicated premises such as \textit{The dimensions of bony and mucosal components of the middle turbinate remain unchanged after septoplasty} which require a stronger reasoning model; besides it curates a larger number of premises increasing the metareasoning difficulty . %While not superficial, they are  compared to Qwen-3 8b's ...  

\subsection{Ablations}
\label{sec:ablation}
We examine choices we made in premise generation and curation: a) The number of entities was limited to 5 because across all datasets and models the average number of entities is $\approx$ 3.5.  b) Obtaining 10 premise pairs was to use experts more than once for related premises. c) Curated premises require a meta-reasoner (Abl. 1). d) Using a stronger premise from each pair reduces conflicting knowledge  (Abl. 2).    e) premise generation involves performance vs. cost decisions for multiple context-derived experts  vs. a single  and iterative  vs. non-iterative  (Abl. 3: Context Priming).

\begin{figure}[t]
    \centering     
    \includegraphics[trim=0in 0in 0 0in,  width=.9 \columnwidth]
     {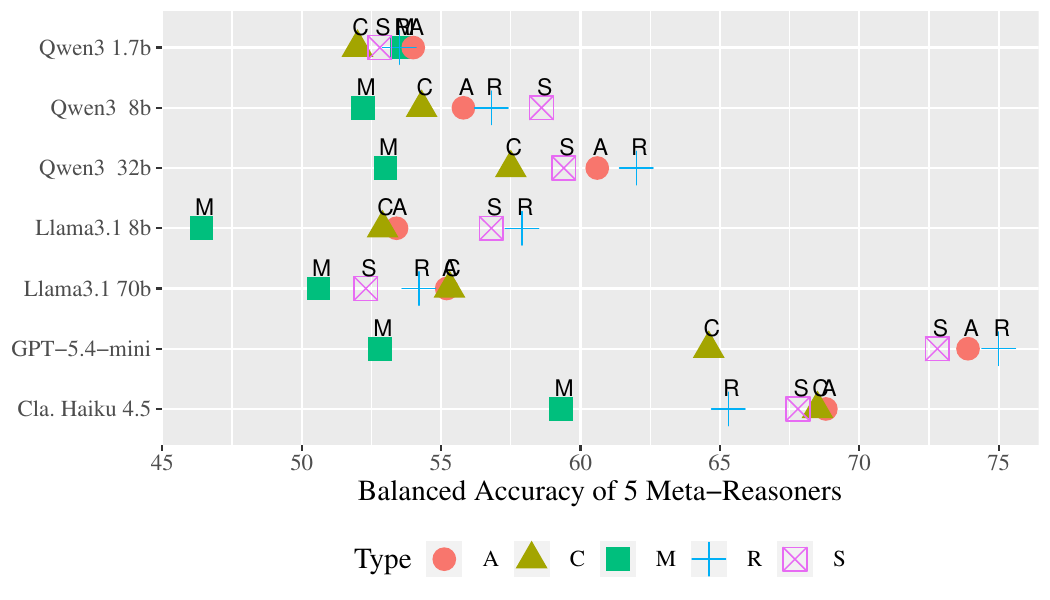}  
     
    \caption{We compare the 5 meta-reasonsers for \textbf{SFBench} claims: M-Majority Voting, R-Reasoning Judge, A-Aggregating Judge, C-Critical Judge and S-Supporting Judge.  Reasoning Judge metareasoner is more consistent across models and datasets and improves performance, by 8\% on SFBench and 5\% on SciFact  (See \cref{fig:metareasoner-performance-scifact} for SciFact and results in Appendix \cref{apptab:verification-options}).  } 
	\label{fig:metareasoner-performance-sfbench}
 \vspace{-2mm}
\end{figure}

\begin{figure}[t]
    \centering     
    \includegraphics[trim=0in 0in 0 0in,  width=.9 \columnwidth]
     {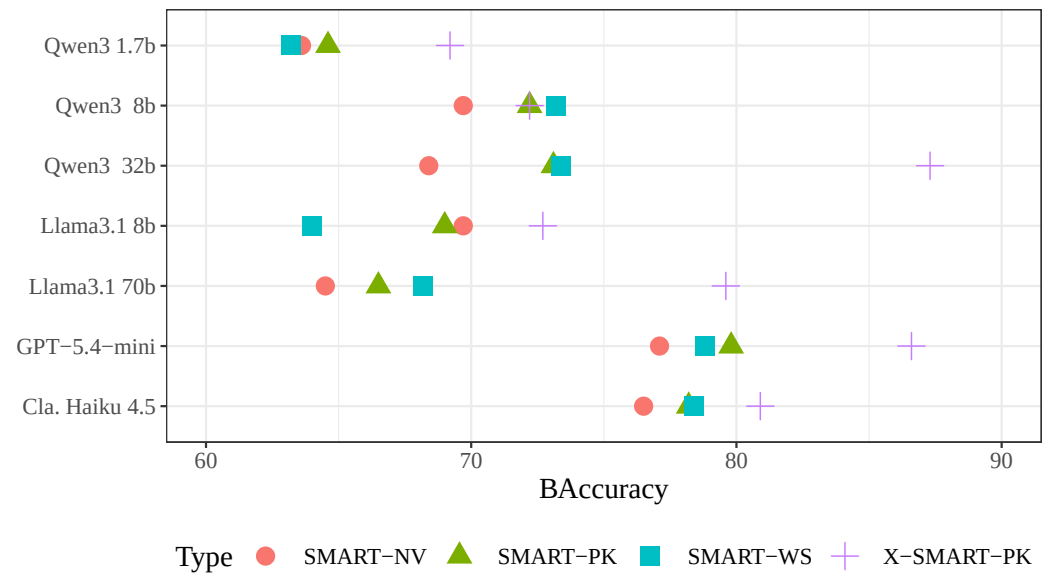} 
    \caption{Comparing SMART-PK, where premises are verified with parametric knowledge, with SMART-NV where no verification is performed on \textbf{SFBench} claims, we show that verification improves performance an average of 2\% with a few exceptions (also true for SciFact claims in \cref{fig:verification-scifact}). Results are in  appendix  \cref{apptab:metareason-options}.     } 
    \label{fig:verification-sfbench}
\end{figure}

\paragraph{Abl. 1 - MetaReasoner: }
As seen in \cref{fig:metareasoner-performance-scifact} and Appendix \cref{fig:metareasoner-performance-sfbench}   the \textbf{Reasoning Judge} consistently performs the best among 4 other meta-reasoners described in \cref{sec:meta-reasoners}. See results in Appendix \cref{apptab:metareason-options}. 

\paragraph{Abl. 2 - Premise Selection:}
 Logically, SMART's use of a stronger premise to avoid information conflicts is sound, however, models are able to reason with both premises as seen in Appendix \cref{fig:smart-premises-both}.

\paragraph{Abl. 3 - Context Priming:}
Appendix \cref{fig:priming-types} compares performance of various premise generation configurations (see Appendix \cref{fig:configurations}) with the \textit{SciFact} collection.  A simple low-resource  method is to sample from top-p premise generations (shown as Non-Iter) using a single static (``science'') expert;  this increases knowledge redundancy (useful for smaller models) but possible for all models:  each API call to the Claude-Haiku-4.5 model provides a single generation and GPT-5.4-mini limits them to 8 (with these models multiple premises can be generated at once, which is not effective for other models). %  useful for Qwen3 1.7b but not the larger models. 
Iterative generation captures fine-grained, related and unique premises, especially, with context-derived experts, improving performance by  2\%.

\section{Conclusions}
\label{sec:conclusions}
 We propose a framework, SMART, to curate solution salient knowledge using a structured prompting method with several LLMs. Our multi-stage setup leads to a natural decomposition of a complex problem and allows multiple agents to reason over simpler problem aspects.  Using a variety of LLMs on 3 complex claim datasets, we demonstrate that knowledge grounded in real-world evidence can   guide reasoning and  improve performance. We show external verification with a noisy websearch and leave improvement of search results for future.

\section{Limitations}
\label{sec:limit}
We acknowledge the pre-trained large language models in our experiments can echo biases and mis-information either implicitly or explicitly. Our work does not alleviate or resolve any of these biases knowingly. 

While we believe our work does not introduce any additional biases, it is likely that our framework may explicate previously hidden implicit biases.  
Our premise generation may draw out implicit biases related to the claims and models are likely to hallucinate premises introducing new biases. Our premise verification is likely to suppress these biases especially when an external source such as the web search is used for verifying premises, unless the bias is present in the source documents as well. 

We believe that a solution should be based on the knowledge at the time of examining the problem.  However, websearch results are not reproducible unless retrieval snapshots or queries are cached. Critical domains will require employing such caching of trusted evidence or maintian domain-specific documents for reproducibility.  Results can also change with slight alterations in prompts, however, as long as we are able to obtain and curate premises we can identify what information led to a particular solution.

\section{Ethical Considerations} 
\label{sec:ethics}

This work analyzes how Large Language Models (LLMs) can be used used for problem-solving. We acknowledge that LLM outputs can reflect biases from pretraining data and we do not make normative claims about models' solutions.   We use  publicly available datasets and papers for this work and these datasets did not contain any personally identifiable information. Additionally, our work does not involve human subjects and does not contain any personally identifiable information. While our work attempts to control for brittleness in models' solution reasoning, it may still produce confident
yet ungrounded solutions for incomplete or under-defined problems.

%\newpage
%\clearpage
\bibliography{anthology,custom}
\bibliographystyle{acl_natbib}
\section{Appendix}
\label{sec:appendix}

No AI assistants were used for writing either the paper or any of the code written for producing the work in this paper.  All writing is original and produced by the authors.

\subsection{SMART Validation with an Example:}
\label{app:data-example}
\prompt[A Scifact-open claim example:]{
\textbf{Claim:} CENP-B inhibits heterochromatin formation

\textbf{Gold Label:} False\\
\\
\textbf{GPT-5-mini’s direct baseline label and reasoning:} True. CENP-B binds alphoid DNA and promotes assembly/maintenance of CENP-A-containing centromeric chromatin and de novo centromere formation; experimental work shows loss of CENP-B impairs CENP-A assembly and can allow heterochromatin features to encroach on centromeric regions, consistent with CENP-B acting to antagonize or inhibit heterochromatin formation at centromeres.\\
\\
\textbf{SMART (using GPT-5.4-mini as agents) generated final label and reasoning:} False. All provided opposing conditions show that CENP-B-containing centromeres bear canonical heterochromatin features (high H3K9me3, low H3K9 demethylase, low transcription, low H3K9ac), which contradicts the claim; published data also link CENP-B with heterochromatin/HP1 recruitment rather than inhibition, so the claim is not supported.\\
\\
\textbf{SMART Framework Stages for the above claim:} For the above final label and reasoning, SMART obtained contrastive premise pairs using GPT-5-mini expert agents. and curated them in the 4 framework stages.\\

\textbf{Identify (Stage 1):} The \cref{appsec:claim-topics-prompt} identified the following 5 entities with modifiable states: CENP-B, histone H3K9 methylation, heterochromatin formation, HP1 recruitment to chromatin, centromeric heterochromatin levels\\
\\
\textbf{Explore (Stage 2):} 5 agents are assigned to be experts, each assigned to be an expert one of the 5 entities obtained above, and prompted to generate the following contrastive premise pairs iteratively using \cref{appsec:claim-explore-prompt}. (We use an ampersand (\&) to separate the two premises in a pair.)\\

1. Presence of CENP-B at centromeres is associated with low levels of the heterochromatin mark H3K9me3 \& Presence of CENP-B at centromeres is associated with high levels of the heterochromatin mark H3K9me3\\
2. HP1 protein levels at centromeric chromatin are low \& HP1 protein levels at centromeric chromatin are high\\
3. Centromeric chromatin is enriched in the histone variant CENP-A. \& Centromeric chromatin is depleted of the histone variant CENP-A.\\
4. CENP-B physically interacts with the histone methyltransferase SUV39H1 \& CENP-B does not physically interact with the histone methyltransferase SUV39H1\\
5. Centromeric DNA contains CENP-B box sequence motifs. \& Centromeric DNA lacks CENP-B box sequence motifs.\\
6. Centromeric chromatin has high levels of the active chromatin mark H3K4me2. \& Centromeric chromatin has low levels of the active chromatin mark H3K4me2.\\
7. Centromeric chromatin has high levels of H3K9-specific demethylase localized at centromeres. \& Centromeric chromatin has low levels of H3K9-specific demethylase localized at centromeres.\\
8. Centromeric transcription levels are high. \& Centromeric transcription levels are low.\\
9. Centromeric chromatin has high levels of H3K9 acetylation (H3K9ac). \& Centromeric chromatin has low levels of H3K9 acetylation (H3K9ac).\\
10. CENP-B is DNA-binding competent \& CENP-B is DNA-binding deficient (to be continued in the next post)\\
\\
\textbf{Verify \& Select (Stage 3):} After verifying the truth value (using \cref{appsec:verify-external-prompt} and identifying the stance of premises (using \cref{appsec:claim-reason-prompt}), the following supporting and opposing premises (each of these oppose the claim being ``valid'') were curated for use in the final reasoning step. No supporting premises were left as the model was more confident about an opposing premise from each pair and assigned it a higher stance rating.  \\

\textbf{Supporting premises:} none \\

\textbf{Opposing premises:} \\
1. Presence of CENP-B at centromeres is associated with high levels of the heterochromatin mark H3K9me3 \textbf{Stance Explanation:} If CENP-B presence at centromeres correlates with high levels of the heterochromatin mark H3K9me3, that contradicts the claim that CENP-B inhibits heterochromatin formation; however, association does not prove causation and contextual factors could confound the relationship, so confidence is high but not absolute.\\
2. Centromeric chromatin has low levels of H3K9-specific demethylase localized at centromeres. \textbf{Stance Explanation:} Low levels of H3K9-specific demethylase at centromeres would favor persistence of H3K9 methylation and heterochromatin formation, which contradicts the claim that CENP-B inhibits heterochromatin; although the condition does not mention CENP-B directly, it is inconsistent with an inhibition role, hence opposition with moderately high confidence.\\
3. Centromeric transcription levels are low. \textbf{Stance Explanation:} Low centromeric transcription is indicative of a heterochromatic, transcriptionally repressed state; this observation contradicts the claim that CENP-B inhibits heterochromatin formation because inhibition of heterochromatin would be expected to raise transcription. While other factors could cause low transcription independent of CENP-B, the condition primarily opposes the claim. \\
4. Centromeric chromatin has low levels of H3K9 acetylation (H3K9ac). \textbf{Stance Explanation:} Low H3K9 acetylation is a hallmark of heterochromatin (acetylation is associated with active/euchromatin), so centromeric chromatin showing low H3K9ac indicates heterochromatin is present at centromeres and thus contradicts the claim that CENP-B inhibits heterochromatin formation; however, other factors could modulate local chromatin so confidence is not absolute.\\
\\
\textbf{Infer (Stage 4):} Using all the supporting and opposing premises the reasoning judge generated the above provided correct label and reasoning using \cref{appsec:claim-reasonj-prompt}.

}

\subsection{Dataset Details}
\label{app:dataset}

\paragraph{\textbf{SFBench\footnote{https://github.com/JHUAPL/SFBench}~\cite{costello2026sfbenchscifyscientificfeasibility}}}is a collection of novel material science claims manually created by subject matter specialists with 5 label categories using a Likert score from -2 to 2. We use all claims with a positive score as `Supported' claims and claims with a negative score as  `Refuted'  claims leading to 183 claims.  We do not always have easy to use textual evidence for these claims, as a result we do not have gold evidence for these claims and we do not evaluate the quality of premises for this dataset.

\prompt[SFBench Data Example:]{
\textbf{Claim:} Lithium manganese iron phosphate can be coated on a carbon cloth current collector at a loading of 3 mAh/cm2 and a formulation of 94/3/3 active/carbon/PVDF binder and flexed repeatedly without cracking or delamination prior to calendering.   \\
\textbf{Likert Score: -1}\\
\textbf{Label:} Contradict
}
\paragraph{SciFact\footnote{https://
github.com/allenai/scifact}~\cite{wadden-etal-2020-fact} \& SciFact-Open\footnote{https://github.com/dwadden/scifact-open}~\cite{wadden-etal-2022-scifact}:} SciFact is a collection of 1409 manual re-writes of citation sentences sampled from well-regarded scientific journals covering basic sciences to clinical medicine. The dataset consists of a dev datasplit (referred to as \textit{SciFact}) and a test datasplit (referred to as \textit{SciFact-Open}).  In \textit{SciFact}, claim labels were manually identified after verifying each claim against the abstract  
of a cited article.  Since the labels are based on a single evidence document which is derived from a small corpus of 5K documents there can be label bias. We use the claims where labels are either `support' or `contradict' for a total of 188 claims. We use the abstract on which the label is based as gold evidence for evaluating our premises.  To reduce  label bias, \textit{SciFact-Open} obtained more evidence documents, a corpus of 50K documents, using a BM25 search and automatically generated labels from SOTA models using all matched evidence documents.  Correlating these automatically generated labels with manually identified labels from the smaller corpus label bias is reduced.  A label from each matched evidence document can lead to several labels for a claim which we consolidate to obtain 206 claims that are exclusively supported or refuted (we drop 73 claims that are both supported and refuted).  We concatenate all the evidence abstracts for our gold evidence which use for evaluating the quality of premises. 

\prompt[SciFact Data Example:]{
\textbf{Claim:} Pseudogene PTENP1 regulates the expression of PTEN by functioning as an miRNA decoy.\\
\textbf{Label:} Support\\
\textbf{Evidence:} A coding-independent function of gene and pseudogene mRNAs regulates tumour biology The canonical role of messenger RNA (mRNA) is to deliver protein-coding information to sites of protein synthesis.  ... 
}

\prompt[SciFact-Open Data Example:]{
\textbf{Claim:} IL-10 production by monocytes encourages CD4 + T cell response.\\
\textbf{Label:} Contradict\\
\textbf{Evidence Docs:} 3 \\ 
\textbf{Evidence:} 1. Programmed death-1-induced interleukin-10 production by monocytes impairs CD4+ T cell activation during HIV infection ... \\
2. Inhibitory and Stimulatory Effects of IL-10 on Human CD8+ T Cells IL-10 is a well-documented ... \\
3. Cutting Edge: Regulatory T cells prevent efficient clearance of Mycobacterium tuberculosis. Mycobacterium tuberculosis remains one of the top microbial killers of humans causing approximately 2 million deaths annually. ...
}

\subsection{Implementation Details \& Prompts}
In all the prompts we use `condition' when refering to a  `premise.'  It was easier to use the term `premise' in the paper after we obtained the results.      

\subsubsection{Prompt Guidelines \& Hyperparameters}
\label{appsec:prompts-guidelines}

\subsection{Infrastructure} We used
both RTX 8000 with 48GB of GPU memory and
Nvidia A100 with 80GB GPU memory for inference with the open models. For the open models we use OpenAI's API endpoint and for Claude we use LiteLLM. Inference time for each claim depends on the hardware used but ranges from 1-5 minutes with Qwen32b and Llama70b taking the longest time. 

\subsection{Websearch}
\label{app:ws}
We use LangChain's implementation of DuckDuckGo Search for obtaining search documents. We use each premise as the query to obtain relevant documents without any date restriction.   With use the text from the top 3 documents along with the url (we do not scrape data from the url) as searched evidence. Using this information as evidence, we verify the premise's truth. When no results are returned are returned, the claim the premise cannot be verified and is not used towards final reasoning. When the results are not relevant the premise is likely to be identified as false and will not be used in the final reasoning. When none of the premises are verified to be true, the final solution falls back to the direct reasoning baseline.  See \cref{apptab:inference-costs}  for an average number of fall-backs (averaged across the 6 datasets).  
We don not cache the evidence across claims or across the 3 runs since its main purpose is to verify a condition before selecting it in final reasoning.  

\subsection{Hyperparameters \& Premise Diversity}
 For simplicity, we use the same settings for all the models and datasets.  We use nucleus sampling to get the top 10\% of generations. These are closely related to the claim from which we obtain multiple sufficiently diverse conditions through repeated sampling. 
We provide a list of guidelines to follow as part of each prompt, such as, not generating duplicate premise pairs, considering the specificity and credibility of any numbers or percentages, assessing any biases and limitations, etc (see \cref{appsec:prompt-common-info}).   
With the Llama models, we encourage premise diversity using a temperature of 0.9, but for all other reasoning we use 0.6 a recommended value for Llama models.  We use the default temperature for Qwen3, GPT-5.4-mini and Claude models.

\subsubsection{Common Guidelines and Output}
\label{appsec:prompt-common-info}
\prompt[Common Guidelines]{
    - Provide a label decision of Y if the claim is supported and N if it is not supported.\\
    - Provide a rating between 0 and 10 to indicate your confidence in the label decision, with 10 indicating a high confidence and 0 indicating no confidence.\\
    - Provide a clear reasoning for why the provided label decision and rating are correct.\\
    - Do not include any additional commentary or keys.
}
\prompt[Prompt Output Format]{
    <example\_format>\\
    \{\\
          "decision": "Y, or N to indicate whether you think the claim is supported or not.",\\
          "rating": "Your rating indicating the confidence in the above label decision on a scale from 0 to 10, 10 indicating high confidence and 0 indicating no confidence.",\\
          "reasoning": "Your brief explanation in one or two sentences for your reasoning on the provided label decision and rating."\\
    \}\\
    </example\_format>
}
\subsubsection{Direct Prompt}
\label{appsec:claim-direct-prompt}
\prompt[Direct Reasoning Prompt]{
        \textbf{User Prompt:}
You are an AI agent designed to assess whether a scientific claim is supported by current scientific knowledge.  The claim is given in the <claim></claim> tags.  You must think and reason carefully, and provide output according to a certain format following the guidelines in the <guidelines></guidelines> tags.
    \\ \\
        Claim: \\
        <claim> \\
    \promptvar{claim context} \\
        </claim> \\
    \\
    Guidelines:\\
        <guidelines>\\
       1. Output a JSON object as shown in <example\_format></example\_format> with exactly three keys: ``decision,'' ``rating'' and ``reasoning''. \\
    \promptvar{common\_guidelines} \\
        </guidelines>\\

    \promptvar{example\_format} \\
    \\ 
        Now, please evaluate the claim above. 
}

\subsubsection{CoT and RAG Prompts}
\label{appsec:claim-cot}
\prompt[CoT Reasoning Prompt]{
        \textbf{User Prompt:}
You are an AI agent designed to assess whether a scientific claim is supported by current scientific knowledge.  The claim is given in the <claim></claim> tags.  You must think step by step and reason carefully, and provide output according to a certain format following the guidelines in the <guidelines></guidelines> tags.
    \\ \\
        Claim: \\
        <claim> \\
    \promptvar{claim context} \\
        </claim> \\
    \\
    Guidelines:\\
        <guidelines>\\
       1. Output a JSON object as shown in <example\_format></example\_format> with exactly three keys: ``decision,'' ``rating'' and ``reasoning''. \\
    \promptvar{common\_guidelines} \\
        </guidelines>\\

    \promptvar{example\_format} \\
    \\ 
        Now, please evaluate the claim above. 
}

\prompt[RAG Reasoning Prompt]{
        \textbf{User Prompt:}
You are an AI agent designed to assess whether a scientific claim is supported by the given evidence.  The claim is given in the <claim></claim> tags and evidence for it is given in the <evidence></evidence> tags.  You must think and reason carefully with the evidence, and provide output according to a certain format following the guidelines in the <guidelines></guidelines> tags.
    \\ \\
        Claim: \\
        <claim> \\
    \promptvar{claim context} \\
        </claim> \\
    \\
    Evidence:\\
        <evidence>\\
     \promptvar{search evidence} \\
        </evidence>
    \\
    Guidelines:\\
        <guidelines>\\
       1. Output a JSON object as shown in <example\_format></example\_format> with exactly three keys: ``decision,'' ``rating'' and ``reasoning''. \\
    \promptvar{common\_guidelines} \\
        </guidelines>\\

    \promptvar{example\_format} \\
    \\ 
        Now, please evaluate the claim above. 
}

\subsubsection{Identifying Salient Entities}
\label{appsec:claim-topics-prompt}
\prompt[]{
        \textbf{System Prompt:} You are an expert at identifying salient entities from a claim's context. 
        \\ \\
        \textbf{User Prompt:}
       You are given a claim in the <claim></claim> tags. Your job is to examine the claim and identify salient entities whose state is modifiable.  If there are more than 5 such entities select the 5 most important entities whose state is most salient to the claim context.  Then, generate these entities separated by a semicolon. You should also follow the guidelines provided in the <guidelines></guidelines> tags. \\
       \\ \\
        Claim: \\
        <claim> \\
        \promptvar{claim} \\
        </claim>
        \\ \\
        Guidelines: \\
        <guidelines> \\
        1. Identify entities from the claim context and consider whether their state is modifiable.\\
        2. Select upto 5 entities whose state is most salient to the claim context.\\
        3. You are allowed to use your internal scientific and linguistic knowledge.\\
        4. Do not list numbers or units by themselves without any associated modifiable state or entity.\\
        5. Do not number them or include any additional commentary or keys.\\
        6. Do not generate any other commentary or preamble, generate only the list of entities. \\
        </guidelines>\\
        \\ 
        Now, please generate a list of entities.
}
\subsubsection{Claim Explore Prompt}
\label{appsec:claim-explore-prompt}
\prompt[]{
        \textbf{System Prompt:} You are an expert with knowledge of properties, states, constraints, rules and laws relating to \promptvar{entity} and have the ability to generate salient contrastive conditions using this knowledge. 
        \\ \\
        \textbf{User Prompt:}     
  You are given a claim in the <claim></claim> tags. You are an expert with knowledge of properties, states, constraints, rules and laws relating to \promptvar{entity}.  With your expert knowledge of \promptvar{entity}, your job is to examine the claim and identify a pair of contrastive reasons or conditions, such as entity properties, states, constraints, rules and laws, that are salient to the claim context and its underlying state.  These conditions should be such that they examine an aspect of the claim to identify if the claim is supported by current scientific knowledge. You should identify contrastive condition pairs that were not previously identified which  are listed in the <identified\_conditions></identified\_conditions> tags. When you are identifying the first condition pair, the <identified\_conditions></identified\_conditions> tags will be empty.  Each pair previously identified is given in the <contrastive\_pair></contrastive\_pair> tags and consists of an expert name in the <expert></expert> tags and the two conditions identified by the expert in the <condition1></condition1> tags and the <condition2></condition2> tags. Pay attention to which expert identified the conditions in a pair to know their viewpoint on the aspect they identified in the conditions. You should provide your output according to a certain format following the provided guidelines in the <guidelines></guidelines> tags. 
        \\ \\
        Claim:
        <claim> \\
         \promptvar{claim}\\
        </claim>\\
        \\ \\
        Identified Conditions: \\
        <identified\_conditions> \\
         \promptvar{<agent>agent</agent>} \\
         \promptvar{<condition1></condition1>} \\
         \promptvar{<condition2></condition2>} \\
        </identified\_conditions> 
        \\ \\
        Guidelines: \\
        <guidelines> \\
        1. Generate a pair of contrastive conditions that are helpful in identifying whether the claim is supported. \\
        2. You are allowed to use your internal knowledge but ensure the conditions are not stated or paraphrased in the claim context.\\
        3. Make sure the conditions in the pair are contrastive. \\
        4. Make sure the condition pair is salient to the claim context. \\
        5. Identify conditions that are related to the previously identified conditions listed in <identified\_conditions></identified\_conditions>. \\
        6. Do not repeat any of the conditions listed in <identified\_conditions></identified\_conditions>. \\
        7. The conditions should be stative expressions.  Do not describe activities. \\
        8. Output your response strictly as a JSON object as shown in <example\_format></example\_format> with exactly two keys: "condition" and "counterfactual". Do not include any additional commentary or keys.\\
        </guidelines>
        \\ \\
        Example output format:\\
        <example\_format> \\
        \{ \\
          "condition": "A property, state, attribute, constraint, rule or law salient to the claim context.",\\
          "counterfactual": "A property, state, attribute, constraint, rule or law contrastive to the condition."\\
        \}\\
        </example\_format>
        \\ \\
        Now, please generate a condition pair for the above claim.
}

\subsubsection{Verify with Parametric Knowledge}
\label{appsec:verify-parametric-prompt}
\prompt[]{
        \textbf{System Prompt:} You are an expert with knowledge of properties, states, constraints, rules and laws and can identify if a given property, state, constraint, rule or law is true.
        \\ \\
        \textbf{User Prompt:}
       You are given a condition in the <condition></condition> tags. The condition describes a property, state, constraint, rule or law governing an entity. Your job is to identify whether the condition is true using your general scientific knowledge. You should also follow the provided guidelines in the <guidelines></guidelines> tags. \\
\\
        Condition: \\
        <condition>\\
     \promptvar{condition}\\
        </condition>\\
\\
        Guidelines:\\
        <guidelines>\\
        1. Evaluate whether the condition is true using your general scientific knowledge. \\
        2. Output your response strictly as a JSON object as shown in <example\_format></example\_format> with exactly three keys: ``decision,'' ``rating'' and ``reasoning''. \\
        3. Provide a label decision of Y if the condition is true and N if it is known to be untrue or it is not likely to be true.\\
        4. Provide a rating between 0 and 10 to indicate your confidence in the label decision, with 10 indicating a high confidence and 0 indicating no confidence.\\
        5. Provide a clear reasoning for why the provided label decision and rating are correct.\\
        6. Do not include any additional commentary or keys.\\ 
        </guidelines>\\
    \\
        Example output format:\\
        <example\_format>\\
       \{\\
          "decision": "Y or N to indicate if the condition is true.",\\
          "rating": "Your rating indicating the confidence in the above label decision on a scale from 0 to 10, 10 indicating high confidence and 0 indicating no confidence.",\\
          "reasoning": "Your brief explanation in one or two sentences for your reasoning on the provided label decision and rating."\\
       \}\\
        </example\_format>\\
    \\
        Now, please evaluate the condition pair above.
}

\subsubsection{Verify with External Evidence}
This uses websearched documents as evidence.
\label{appsec:verify-external-prompt}
\prompt[]{
        \textbf{System Prompt:}  You are an expert with knowledge of properties, states, constraints, rules and laws and can identify if a given property, state, constraint, rule or law is true.
        \\ \\
        \textbf{User Prompt:}
        You are given a condition in the <condition></condition> tags and some evidence in the <evidence></evidence> tags. The condition describes a property, state, constraint, rule or law governing an entity. Your job is to identify whether the condition is true based on the evidence. You should also follow the provided guidelines in the <guidelines></guidelines> tags. \\
        \\ \\
        Condition: \\
        <condition> \\
    \promptvar{condition}  \\
        </condition> 
        \\ \\
        Evidence: \\
        <evidence> \\
    \promptvar{evidence}  \\
        </evidence> \\
        \\ \\
        Guidelines: \\
        <guidelines> \\
        1. Analyze whether the evidence provides sufficient context and scope for verifying whether the condition is true.\\
        2. Evaluate whether the condition is true based on the provided evidence and your scientific knowledge. \\ %If the evidence does not provide any useful information  use your scientific knowledge.\\
        3. Output your response strictly as a JSON object as shown in <example\_format></example\_format> with exactly three keys: ``decision,'' ``rating'' and ``reasoning''. \\
        4. Provide a label decision of Y if the condition is true and N if it is known to be untrue or it is not likely to be true based on the provided evidence and your scientific knowledge.\\
        5. Provide a rating between 0 and 10 to indicate your confidence in the label decision, with 10 indicating a high confidence and 0 indicating no confidence.\\
        6. Provide a clear reasoning for why the provided label decision and rating are correct.\\
        7. Do not include any additional commentary or keys.\\ 
        </guidelines>\\
    \\
        Example output format:\\
        <example\_format>\\
        \{\\
          "decision": "Y or N to indicate if the condition is true.",\\
          "rating": "Your rating indicating the confidence in the above label decision on a scale from 0 to 10, 10 indicating high confidence and 0 indicating no confidence.",\\
          "reasoning": "Your brief explanation in one or two sentences for your reasoning on the provided label decision and rating."\\
        \}\\
        </example\_format>\\
    \\
        Now, please evaluate the condition pair above.

}

\subsubsection{Obtaining a Premise's Stance}
\label{appsec:claim-reason-prompt}
\prompt[]{
        \textbf{System Prompt:} 
        You are an expert at reasoning about conditions that support or oppose claims related to medical and science topics.
        \\ \\
        \textbf{User Prompt:}
        You are given a claim in the <claim></claim> tags and a condition <condition></condition> tags. Your job is to analyze the claim and the condition and decide whether the condition supports or opposes the claim using your general scientific knowledge. You should also follow the provided guidelines in the <guidelines></guidelines> tags. 
        \\ \\
        Claim:  \\
        <claim>  \\
    \promptvar{claim}  \\
        </claim>
        \\ \\        
        Condition: \\
        <condition> \\
    \promptvar{condition} \\
        </condition>
        \\ \\
        Guidelines:  \\
        <guidelines> \\
        1. Analyze the context and scope of the claim.\\
        2. Consider the specificity and credibility of any numbers or percentages.\\
        3. Assess any potential biases or limitations.\\
        4. Evaluate using your general scientific knowledge whether the condition supports or opposes the claim. \\
        5. It is likely the condition both supports and opposes the claim on different aspects.  For such a condition use the unsure label.\\
        6. It is likely the condition is unrelated to the claim or related to the claim but neutral towards it.  For such a condition use the unsure label.\\
        7. Output your response strictly as a JSON object as shown in <example\_format></example\_format> with exactly three keys: ``decision,'' ``rating'' and ``reasoning''. \\
        8. Provide a label decision of Y if the condition supports the claim, N if the condition opposes the claim, and U for all other situations.\\
        9. Provide a rating between 0 and 10 to indicate your confidence in the label decision, with 10 indicating a high confidence and 0 indicating no confidence.\\
        10. Provide a clear reasoning for why the provided label decision and rating are correct.\\
        11. Do not include any additional commentary or keys. \\
        </guidelines>\\
    \\
        Example output format:\\
        <example\_format>\\
        \{\\
          "decision": "Y or N or U to indicate the condition's support for the claim.",\\
          "rating": "Your rating indicating the confidence in the above label decision on a scale from 0 to 10, 10 indicating high confidence and 0 indicating no confidence.",\\
          "reasoning": "Your brief explanation in one or two sentences for your reasoning on the provided label decision and rating."\\
        \}\\
        </example\_format>\\
         \\
        Now, please evaluate the claim and condition pair above. 
}

\subsubsection{Claim Aggregate Explanations Prompt}
\label{appsec:claim-aggexp-prompt}
\prompt[]{
        \textbf{System Prompt:} You are an expert at summarizing reasons given for whether claims about medical and science topics are true. 
        \\ \\
        \textbf{User Prompt:}
        You are given a claim in the <claim></claim> tags and multiple conditions that support the claim in the <supporting\_conditions></supporting\_conditions> and multiple conditions that oppose the claim in the <opposing\_conditions></opposing\_conditions> tags. Each condition in the <condition></condition> tags is verified for its truth value and whether it supports or opposes the claim feasibility in the <explanation></explanation> tags. Summarize the justifications provided in all the <explanation></explanation> tags. The <supporting\_conditions></supporting\_conditions> tags or the <opposing\_conditions></opposing\_conditions> tags may be empty and in such a case state that no conditions were provided.
        \\ \\
        Claim:\\
        <claim>\\
    \promptvar{claim}  \\
        </claim>
        \\ \\
        Claim Supporting Conditions:\\
        <supporting\_conditions>\\
        \promptvar{<condition></condition><explanation></explanation>}\\
        \promptvar{<condition></condition><explanation></explanation>}\\
        \promptvar{...} \\
        </supporting\_conditions>
        \\ \\
        Claim Opposing Conditions:\\
        <opposing\_conditions>\\
        \promptvar{<condition></condition><explanation></explanation>}\\
        \promptvar{<condition></condition><explanation></explanation>}\\
        \promptvar{...} \\
        </opposing\_conditions>\\
        \\ 
        Now, please summarize the conditions above.
}

\subsubsection{Claim Aggregate Judge Prompt}
\label{appsec:claim-aggregatej-prompt}
\prompt[]{
        \textbf{System Prompt:}  You are an expert at critically examining and deciding whether claims about medical and science topics are true.
        \\ \\
        \textbf{User Prompt:}
  You are a scientific analyst designed to assess whether a scientific claim is supported by current scientific knowledge. The claim is given in the <claim></claim> tags. You must think and reason logically using the additional information provided as supporting and opposing conditions to aid in your reasoning. Conditions that support the claim are given in the <supporting\_conditions></supporting\_conditions> tags and conditions that oppose the claim are given in the <opposing\_conditions></opposing\_conditions> tags. Each of the conditions is given in the <condition></condition> tags and their justification for supporting or opposing the claim is given in the <explanation></explanation> tags.  Use these conditions to aid in your reasoning about whether the claim is supported. Provide your output according to a certain format following the guidelines in the <guidelines></guidelines> tags.
\\ \\
        Claim:\\
        <claim>\\
    \promptvar{claim}  \\
        </claim>
        \\ \\
        Claim Supporting Conditions:\\
        <supporting\_conditions>\\
        \promptvar{<condition></condition><explanation></explanation>}\\
        \promptvar{<condition></condition><explanation></explanation>}\\
        \promptvar{...} \\
        </supporting\_conditions>
\\ \\
        Claim Opposing Conditions:\\
        <opposing\_conditions>\\
        \promptvar{<condition></condition><explanation></explanation>}\\
        \promptvar{<condition></condition><explanation></explanation>}\\
        \promptvar{...} \\
        </opposing\_conditions>
        \\ \\
        Guidelines:\\
        <guidelines>\\
        1. Analyze the claim using your scientific knowledge.\\
        2. Consider the specificity and credibility of any numbers or percentages.\\
        3. Assess any potential biases or limitations.\\
        4. Examine all the supporting and opposing conditions and their explanations. Use these to aid your scientific knowledge and reason logically about whether the claim is supported.\\
        5. Output a JSON object as shown in <example\_format></example\_format> with exactly three keys: ``decision,'' ``rating'' and ``reasoning''.\\
         \promptvar{common\_guidelines} \\
        </guidelines>\\

    \promptvar{example\_format} 
    \\ 
       \\
        Now, please evaluate the claim above. 

}

\subsubsection{Claim Reasoning Judge Prompt}
\label{appsec:claim-reasonj-prompt}
\prompt[]{
        \textbf{System Prompt:} You are an expert at critically examining and deciding whether claims about medical and science topics are true. 
        \\ \\
        \textbf{User Prompt:}
      You are a scientific analyst designed to assess whether a scientific claim is supported by current scientific knowledge. The claim is given in the <claim><\/claim> tags. You must think and reason logically using the additional information provided as supporting and opposing conditions to aid in your reasoning. Conditions that support the claim are given in the <supporting\_conditions></supporting\_conditions> tags and conditions that oppose the claim are given in the <opposing\_conditions></opposing\_conditions> tags. Each of the conditions is given in the <condition></condition> tags.  Use these conditions to aid in your reasoning about whether the claim is supported. Provide your output according to a certain format following the guidelines in the <guidelines></guidelines> tags.
\\ \\
        Claim:\\
        <claim>\\
    \promptvar{claim}  \\
        </claim>
        \\ \\
        Claim Supporting Conditions:\\
        <supporting\_conditions>\\
        \promptvar{<condition></condition>} \\
        \promptvar{<condition></condition>}\\
        \promptvar{...} \\
        </supporting\_conditions>
\\ \\
        Claim Opposing Conditions:\\
        <opposing\_conditions>\\
        \promptvar{<condition></condition>} \\
        \promptvar{<condition></condition>}\\
        \promptvar{...} \\
        </opposing\_conditions>
        \\ \\
        Guidelines:\\
        <guidelines>\\
        1. Analyze the claim using your scientific knowledge.\\
        2. Consider the specificity and credibility of any numbers or percentages.\\
        3. Assess any potential biases or limitations.\\
        4. Examine all the supporting and opposing conditions and use these to aid your scientific knowledge and reason logically about whether the claim is supported.\\
        5. Output a JSON object as shown in <example\_format></example\_format> with exactly three keys: ``decision,'' ``rating'' and ``reasoning''. \\
        \promptvar{common\_guidelines} \\
        </guidelines>\\

    \promptvar{example\_format}  \\
       \\
        Now, please evaluate the claim above. 
}

\subsubsection{Claim Critical Judge Prompt}
\label{appsec:claim-criticalj-prompt}
\prompt[]{
        \textbf{System Prompt:} You are an expert at critically examining and deciding whether claims about medical and science topics are true.
        \\ \\
        \textbf{User Prompt:}
     You are a scientific analyst designed to assess whether a scientific claim is supported by current scientific knowledge. The claim is given in the <claim></claim> tags. You must think and reason logically using the additional information provided as opposing conditions to aid in your reasoning. Conditions that oppose the claim are given in the <opposing\_conditions></opposing\_conditions> tags where each condition is given in the <condition></condition> tags.  Use these conditions to play a devil's advocate and interrogate your knowledge relating to the claim. Provide your output according to a certain format following the guidelines in the <guidelines></guidelines> tags.
\\ \\
        Claim:\\
        <claim>\\
        \promptvar{claim} \\
        </claim>
        \\ \\
        Claim Opposing Conditions:\\
        <opposing\_conditions>\\
        \promptvar{<condition></condition>} \\
        \promptvar{<condition></condition>}\\
        \promptvar{...} \\
        </opposing\_conditions>\\
         \\
        Guidelines:\\
        <guidelines>\\
        1. Analyze the claim using your scientific knowledge.\\
        2. Consider the specificity and credibility of any numbers or percentages.\\
        3. Assess any potential biases or limitations.\\
        4. Examine all the opposing conditions and use them to critically reason about whether the claim is supported taking the stance of a devil's advocate.\\
        5. Output a JSON object as shown in <example\_format></example\_format> with exactly three keys: ``decision,'' ``rating'' and ``reasoning''. \\
        \promptvar{common\_guidelines} \\
        </guidelines>\\

    \promptvar{example\_format} \\
    \\ 
    Now, please evaluate the claim above. 

}

\subsubsection{Claim Supportive Judge Prompt}
\label{appsec:claim-supportivej-prompt}
\prompt[]{
        \textbf{System Prompt:} You are an expert at critically examining and deciding whether claims about medical and science topics are true.
        \\ \\
        \textbf{User Prompt:}
            You are a scientific analyst designed to assess whether a scientific claim is supported by current scientific knowledge. The claim is given in the <claim></claim> tags. You must think and reason logically using the additional information provided as supporting conditions to aid in your reasoning. Conditions that support the claim are given in the <supporting\_conditions></supporting\_conditions> tags where each condition is given in the <condition></condition> tags.  Use these conditions to examine for contradictory information that prevents these these conditions from supporting the claim. Provide your output according to a certain format following the guidelines in the <guidelines></guidelines> tags. 
\\ \\
        Claim:\\
        <claim>\\
        \promptvar{claim} \\
        </claim>
        \\ \\
        Claim Supporting Conditions:\\
        <supporting\_conditions>\\
        \promptvar{<condition></condition>} \\
        \promptvar{<condition></condition>}\\
        \promptvar{...} \\
        </supporting\_conditions>
        \\ \\
        Guidelines:\\
        <guidelines>\\
        1. Analyze the claim using your scientific knowledge.\\
        2. Consider the specificity and credibility of any numbers or percentages.\\
        3. Assess any potential biases or limitations.\\
        4. Examine all the supporting conditions and use them to critically reason about whether the claim is supported.\\
        5. Output a JSON object as shown in <example\_format></example\_format> with exactly three keys: ``decision,'' ``rating'' and ``reasoning''. \\
        \promptvar{common\_guidelines} \\
        </guidelines>\\

    \promptvar{example\_format} \\
       \\
        Now, please evaluate the claim above. 

}

\subsubsection{Refute Filter}
\label{appsec:refute-filter-prompt}
\prompt[]{
        \textbf{System Prompt:} You are an expert at evaluating rationales.
        \\ \\
        \textbf{User Prompt:}
        You are asked to assess whether the rationale below provides specific details for why the scientific claim is refuted or if it uses lack of evidence for supporting the claim as the reason to refute it. Think and reason carefully and provide your response strictly as a JSON object as shown in the output. \\
        \\
        Claim:      \promptvar{claim}
        \\
        Rationale:    \promptvar{rationale}
        \\
        Output:\\
        \{\\
            "label": "provide Y if the rationale contains specific details, provide N if the rationale uses lack of evidence for supporting the claim as the reason to refute it",\\
            "rationale": "Provide a clear reasoning for the label in one or two sentences"\\
        \}\\
\\
        Now, please assess the rationale.

}

\subsection{Related Work (expanded from main paper section \cref{sec:related-work} )}
\label{app:related}
Below, we outline research related to the ideas we use in our work.
\paragraph{Implicit and Counterfactual States}
Causal relationships between implicit states and events have been previously studied using common everyday events~\cite{tandon-etal-2020-dataset,tandon-etal-2019-wiqa,ghosh-etal-2023-pasta,vallurupalli-ferraro-2025-core}.  Leveraging their idea of contrastive states, we obtain contrastive premises capturing implicit knowledge in the problem context.  

\paragraph{Language Models \& Complex Reasoning}
Complex reasoning requires models to go beyond exploiting surface level data patterns and reason with knowledge embedded in the context.  This requires a model to connect distinct pieces of information and fill-in missing gaps.  \newcite{zhao-etal-2023-complex}'s tutorial covers a variety of methods and past works that examined various aspects of complex reasoning.  We use some of these methods such as leveraging external knowledge to validate conditions and few-shot exemplars to obtain contrastive states.

\paragraph {Iterative and Diverse Reasoning}
Techniques such as Self-Correction~\cite{welleck2023generating}, Self-Refine~\cite{madaan2023selfrefine}, ReAct~\cite{yao_react} and Reflexion~\cite{shinn2023reflexion} use an iterative approach to improve problem output through incorporating feedback from previous reasoning steps.  These methods use a iterative approach to post-hoc correct the solution while we obtain solution-salient knowledge prior to the solution. Techniques such as   DiVeRSe~\cite{li-etal-2023-making}, Tree-of-thought ~\cite{yao_tot}, Graph-of-thought~\cite{besta2024got} examine diverse reasoning paths selecting optimal intermediate solutions.  Our reasoning approach selects optimal premises in lieu of intermediate solutions.  Multi agent debate frameworks leverage iterative and diverse reasoning paths~\cite{du2024improving,liang-etal-2024-encouraging,xiong-etal-2023-examining}, however, ~\newcite{gu-etal-2025-explain} showed that instead of a debate multiple agents can also be leveraged to handle sub-problems. We take inspiration from the latter and decompose obtaining a problem's implicit state into sub-problems.

 \subsection{Solution Inference with other Meta-Reasoners }
 \label{sec:meta-reasoners}
We compare with 4 other meta-reasoners: majority voting, an aggregating judge, a critical judge and a supportive judge.   

The \textbf{Majority Voting} meta-reasoner is arguably the simplest one: each stance label contributes equally to the solution and the solution is based on the higher average stance rating $s^r$ of premises in the two sets: $P_s$ or $P_o$. The claim label $y^l$ is 'support' if the average stance rating of premises in $P_s$ is higher, else it is 'refute.'  The confidence rating for this label $y^r$ is equal to the average $s^r$ of all premises in the set that determined the label with $y^e$ summarizing all premises in $P_s$ \& $P_o$.   

%The \textbf{Reasoning Judge} meta-reasoner \textbf{in the SMART framework} generates $y$ conditioned on the supporting and opposing premises, i.e.,  $\{y |x, p_k \in (P_s, P_o)\}$.  This allows the premises to be interpreted together where model reasoning coherently connects these premises, inherently assigning different weights to them.

The \textbf{Aggregating Judge}, like the reasoning judge, selects $y$ conditioned on all supporting and opposing premises.  It additionally uses all the stance explanations, i.e., $\{(y |x, (p_k,s^e_k) \in (P_s, P_o)\}$,  to provide additional context which can be useful to connect the premises especially when small changes in wording determine the claim's veracity. However, these justifications can bias the final inference, as they are obtained through isolated reasoning with each premise. 

The \textbf{Critical Judge} reasons only with the opposing premises playing the role of ``the devil's advocate,'' critically considering whether each premise can discredit the claim.  The solution is conditioned on opposing premises only, i.e., 
$\{y |x,p_k \in P_o)\}$.

The \textbf{Supportive Judge} reasons with only the supporting premises, considering whether they can support the claim.  The solution is conditioned on supporting premises only,  i.e.,   
$\{y|x,p_k \in P_s)\}$, contrasting the reasoning of the Critical Judge.

\subsection{Issues with the Gemma models}
\label{app:gemma}
We found that Gemma family models underperformed comparably sized models from Llama and Qwen3. This was mainly because these model were unable to generate  a mix of both opposing and supporting premises. Qualitatively, we noticed that Gemma models curate mostly supporting premises which improves the accuracy of valid claims but lowers the accuracy of invalid claims, leading to an overall poorer performance than the CoT baseline.  In addition, CoT performance is boosted as lack of supporting evidence as these models use the lack of evidence to refute claims as seen in \cref{apptab:claim-validation-configD-gemma}. 

\begin{table}[t]
 \centering
\resizebox{.99\columnwidth}{!}{
 \begin{tabular}{|l|c|c|c|}
 \hline
 Model& CoT& SMART-PK &SMART-WS  \\
\hline
Gemma 3 1b    &.563$\pm$.003&.562$\pm$.005&.562$\pm$.005 \\
Gemma 3 4b    &.622$\pm$.001&.594$\pm$.007&.609$\pm$.011 \\
Gemma 3 12b   &.680$\pm$.002&.654$\pm$.003&.651$\pm$.005 \\
\hline
\end{tabular}
}
 \caption{\textbf{Gemma Models} do not make for good SMART agents for two reasons: a) CoT performance is boosted by refuting claims for which there is no supporting  evidence. b) Models are good at generating supporting premises but poor at generating contrastive premises that can lead to refuting evidence. This results in SMART meta-reasoning being overly positive with not having opposing reasons for a claim. As a result, we do not use these models in examining other data besides SciFact. This behavior is also seen to a lesser extent with Llama3.1 70b model.  }
\label{apptab:claim-validation-configD-gemma}
%\vspace{-3mm}
\end{table}

\begin{figure}[t]
    \centering     
    \includegraphics[trim=0in 0in 0 0in,  width=.9 \columnwidth]
    {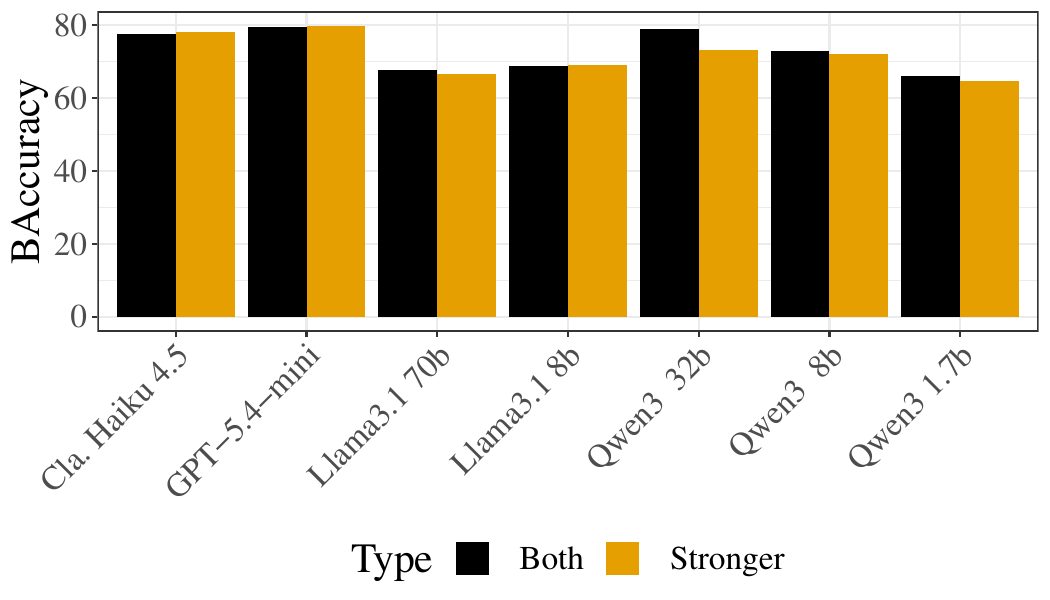} 
        \caption{Comparing SMART-PK performance using a stronger premise Vs. both premises. Logically using a stronger premise was a good decision (performance of both both GPT-5.4-mini and Claude-Haiku-4.5 reduces due to direct conflicts), however, model reasoning is able to use the additional nuanced information provided by both premises. Larger models are better at leveraging the nuanced information in both premises (Llama70b improves by 1.3\% \& Qwen3 32b by 6.7\%); smaller models benefit from additional premises as they do not curate as many premises (Qwen3 1b improves by 1.5\%).  } 
    \label{fig:smart-premises-both}
 %    \vspace{-3mm}
\end{figure}
\begin{figure}[t]
    \centering     
    \includegraphics[trim=0 0.3in 0 0in,  width=.9 \columnwidth]
    {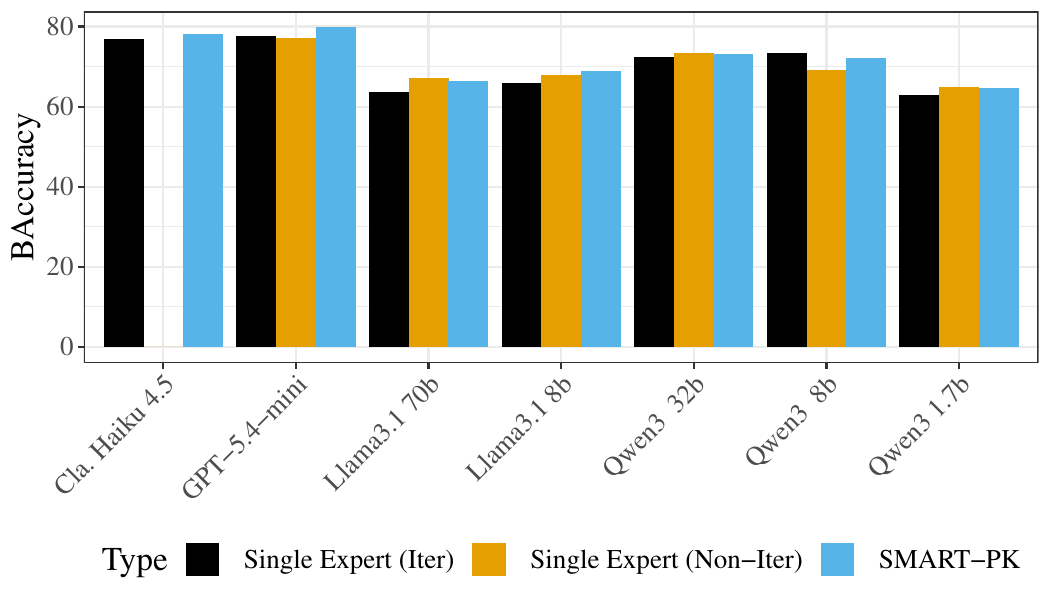} 
    \caption{We compare various premise generation methods using the \textit{SciFact} collection. We obtain premises with a single expert in one-shot (Non-Iter) Vs an iterative generation strategy (Iter). We also compare SMART's multiple context-derived experts with the iterative strategy. Claude-Haiku-4.5 does not allow obtaining more than one generation per API call; due to this we do not have any result for the single expert in one-shot (Non-Iter) strategy. %using few-shot examples (Iter FS) with no examples (Iter ZS). %We verify premises with gold evidence and use the Reasoning Judge meta-reasoner for the final solution.
    } %
	\label{fig:priming-types}
% \vspace{-5mm}
\end{figure}

\begin{figure}[t]
    \centering     
    \includegraphics[trim=0 0.3in 0 0in,  width=.9 \columnwidth]
    {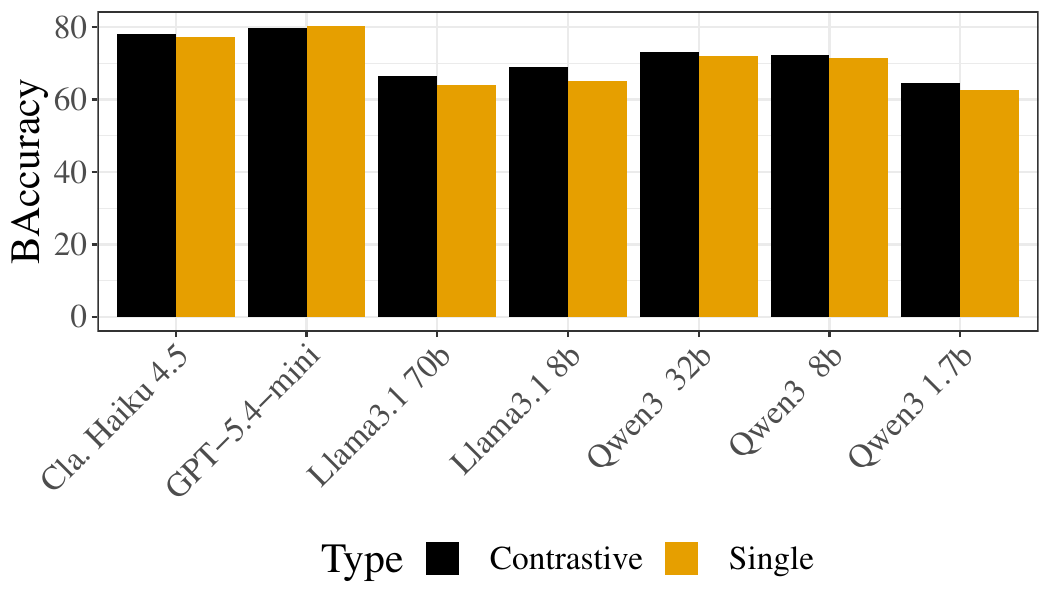} 
    \caption{Comparing SMART-PK performance generating a single premise Vs. contrastive premises on SciFact claims. See  \cref{fig:smart-premises-single-sfbench} for SFBench claims.  Contrastive premises improve performance by 3\% across datasets and models. See results in \cref{apptab:model-single-constrastive-performance} } 
    \label{fig:smart-premises-single-scifact}
     \vspace{-3mm}
\end{figure}

\subsection{Error Types:} 
\label{appsec:error-types}
Examining errors from both claim sources, we find that incomplete model knowledge relating to a claim leads to errors.  Some errors are silent where the model selects the correct label, either due to \textbf{model positivity} or inadequate evidence supporting a \textit{refute label}.  We filter the latter with the \textbf{Refute Filter} but filtering the former positivity errors is not simple as these rationales contain claim knowledge, although, the knowledge is incomplete.) 
We roughly classify these errors into the following types: 

a) When models do not have complete knowledge, CoT ignores unknown aspects and \textit{supports} the claim \textbf{(Model Positivity)}. SMART explores these unknowns with the premise knowledge and either leads to the correct \textit{refuted} label \textbf{[Type 1 (these situations are not errors)]}, or \textit{supports} the claim with the same model positivity due to incomplete premise knowledge \textbf{[Type 3]}.  There are a few Type 1 and a few Type 3 errors across all models.

b) Models have little to no claim related knowledge, yet, CoT correctly identifies with a \textit{refute} label, with the rationale citing lack of support for the claim. \textbf{[Type 2b]}.  Meanwhile, SMART reasoning either correctly reasons with the premise knowledge, or leads to the incorrect label when the knowledge provided by the premises is incomplete (a lack of opposing conditions either due to a model  invalidating them from lack of knowledge or a model not generating them as with Llama3.1 70b are the main reasons for this situation) \textbf{[Type 2a]}.   When lacking curated premises, SMART's Fallback to direct-reasoning leads to the same Type 2b error seen with CoT.

Even though models identify the correct label, Type 2b are also errors due to not providing adequate support for the label.  There are far more of these errors which artificially inflate performance and we handle these with a \textbf{Refute Filter} (see \cref{sec:results}).  This filtering removes a few correct labels, especially seen with GPT-5.4-mini and Claude-Haiku-4.5, where models correctly cite that there are no known reasons to support the claim.  SMART-WS can correct both Types 2a and 2b errors with external evidence helping  curate better premises. Models' performance mainly differs in the number of these errors.
 
d) Both CoT and SMART are incorrect due to  a lack of knowledge.  Similar to a) \textbf{Model Positivity} leads to errors in \textit{refuted} labels. \textbf{[Type 3]} 

Type 1 situations improve SMART performance over CoT; Type 2a \& 2b lower the performance for both CoT and SMART after refute filtering;  Type 3 lower the performance of both.

%\subsection{Additional Results:}
%\label{app:additional-results}
%Due to space constraints we provided links to the following tables and figures in the main paper.

\begin{table}[t]
 \centering
 \resizebox{.99\columnwidth}{!}{
  \begin{tabular}{|l|ccc|ccc|}
 \hline
 Dataset---------$>$  &      \multicolumn{3}{|c|}{SFBench} & \multicolumn{3}{|c|}{SciFact}  \\
Model   & CoT & SMART-PK & SMART-S & CoT & SMART-PK & SMART-S\\ 
   \hline
 Qwen3 1.7b   &56 &43 &58 &78 &46 &48 \\
Qwen3 8b      &28 &20 &53 &63 &31 &24 \\
Qwen3 32b     &12 &09 &26 &67 &32 &11 \\
Llama 8b      &48 &27 &51 &87 &99 &29 \\
Llama 70b     &13 &12 &47 &61 &27 &18 \\
GPT-5.4-mini  & 8 & 7 &24 &23 &16 &13 \\
Cla. Haiku 4.5&22 & 7  &48 &51 &11 &19 \\
\hline
\end{tabular}
}
 \caption{Comparing the percentage of rationales that lack support for a \textit{refute} claim that are filtered by the \textbf{Refute Filter} for CoT reasoning, SMART-PK reasoning with contrastive premises and the single premise generation method (SMART-S).   These lead to incorrect logical reasoning on 50\% of SFBench claims and 40\% of SciFact claims (claims with a gold \textit{refute}). CoT leads to almost double the errors than SMART on smaller models.  Single premise generation leads to more errors on SFBench (double the number than on SciFact) where as the reverse is true for contrastive premises.   We discuss these results in \cref{sec:results}.   
 }
\label{apptab:filtered-refute}
%\vspace{-3mm}
\end{table}
                  
\begin{table*}[t]
 \centering
 \resizebox{.99\textwidth}{!}{
  \begin{tabular}{l|l|ccr||ccr||ccr||c}
 \hline
  &   &      \multicolumn{3}{|c|}{Chain-of-Thought} & \multicolumn{3}{|c}{ RAG with Web Search}  & \multicolumn{3}{|c|}{SMART-PK}  &\%SMART \\
Data & Model & Before RF &  After RF & \textbf{\%Drop$\downarrow$} & Before RF&  After RF & \textbf{\%Drop$\downarrow$}& Before RF & After RF &\textbf{\%Drop$\downarrow$}   &Improv.\\
\hline
&Qwen3 1.7b         &.534$\pm$.000 &.165$\pm$.006 &36.9 &.503$\pm$.008&.100$\pm$.008 &40.3 &.535$\pm$.023 &.274$\pm$.025 &26.1 & 10.9\\
&Qwen3 8b           &.610$\pm$.006 &.369$\pm$.005 &34.1 &.500$\pm$.000&.000$\pm$.000 &50.0 &.568$\pm$.009 &.435$\pm$.016 &13.3 &  6.6\\
&Qwen3 32b          &.642$\pm$.000 &.362$\pm$.005 &28.0 &.507$\pm$.068&.096$\pm$.022 &43.9 &.620$\pm$.019 &.569$\pm$.032 & 5.1 & 20.7\\
SFBench&Llama 8b    &.523$\pm$.000 &.243$\pm$.099 &28.0 &.514$\pm$.009&.038$\pm$.006 &47.6 &.579$\pm$.004 &.447$\pm$.010 &13.2 & 20.4\\
&Llama 70b          &.643$\pm$.000 &.514$\pm$.000 &12.9 &.519$\pm$.003&.040$\pm$.008 &47.9 &.542$\pm$.023 &.536$\pm$.030 & 0.6 &  2.2\\
&GPT-5.4-mini       &.713$\pm$.012 &.600$\pm$.020 &11.3 &.503$\pm$.004&.003$\pm$.004 &50.0 &.750$\pm$.026 &.702$\pm$.004 & 4.8 & 10.2\\
&Cla. Haiku 4.5     &.654$\pm$.004 &.522$\pm$.008 &13.2 &.506$\pm$.016&.011$\pm$.008 &49.5 &.653$\pm$.012 &.583$\pm$.004 & 7.0 &  6.1\\
\hline
&Qwen3 1.7b         &.590$\pm$.000 &.377$\pm$.007 &21.3 &.619$\pm$.027&.240$\pm$.050 &37.9 &.646$\pm$.019 &.448$\pm$.044 &19.8 &  7.1\\
&Qwen3 8b           &.707$\pm$.000 &.548$\pm$.011 &15.9 &.654$\pm$.037&.204$\pm$.065 &45.0 &.722$\pm$.014 &.641$\pm$.007 & 8.1 &  9.3\\
SciFact \&&Qwen3 32b&.652$\pm$.003 &.616$\pm$.009 & 3.6 &.656$\pm$.049&.192$\pm$.065 &46.4 &.731$\pm$.014 &.696$\pm$.018 & 3.5 &  8.0\\
SciFact-&Llama 8b   &.498$\pm$.000 &.035$\pm$.011 &46.3 &.654$\pm$.045&.258$\pm$.083 &39.6 &.690$\pm$.025 &.595$\pm$.015 & 9.5 & 56.0\\
Open&Llama 70b      &.699$\pm$.004 &.630$\pm$.000 & 6.9 &.668$\pm$.057&.306$\pm$.093 &36.2 &.665$\pm$.012 &.654$\pm$.014 & 1.1 &  2.4\\
&GPT-5.4-mini       &.826$\pm$.004 &.750$\pm$.006 & 7.6 &.663$\pm$.004&.202$\pm$.002 &46.1 &.798$\pm$.006 &.765$\pm$.008 & 3.3 &  1.5\\
&Cla. Haiku4.5      &.796$\pm$.003 &.736$\pm$.019 & 6.0 &.589$\pm$.038&.107$\pm$.031 &48.2 &.782$\pm$.014 &.749$\pm$.017 & 3.3 &  1.3\\
%\hline 
%&Qwen3 1.7b           &.563$\pm$.000 &.583$\pm$.014 &2.0 &.545$\pm$.010 &.606$\pm$.008 & 6.1\\
%&Qwen3 8b             &.613$\pm$.000 &.635$\pm$.005 &2.2 &.618$\pm$.082 &.630$\pm$.002 & 1.2\\
%&Qwen3 32b            &.652$\pm$.001 &.662$\pm$.010 &1.0 &.707$\pm$.083 &.645$\pm$.005 & 6.2\\
%AVeriTeC&Llama3.1 8b  &.541$\pm$.000 &.640$\pm$.014 &9.9 &.618$\pm$.092 &.611$\pm$.019 & -0.7\\
%&Llama3.1 70b         &.742$\pm$.008 &.726$\pm$.004 &-1.6 &.623$\pm$.057 &.732$\pm$.008 & 10.9\\
%&GPT-5.4-mini         &.800$\pm$.004 &.784$\pm$.026 &-1.6 &.617$\pm$.061 &.800$\pm$.035 & 18.3\\
%&Cla. Haiku 4.5       &.775$\pm$.002 &.784$\pm$.002 &0.9 &.577$\pm$.037 &.781$\pm$.003 & 20.4\\
\hline
\end{tabular}
}
 \caption{Comparing the balanced accuracy of CoT, RAG and SMART-PK before and after applying the \textbf{Refute Filter} (\textit{refute} labels that are not adequately supported are changed to 'unknown', essentially dropping them from contributing to balanced accuracy). Smaller models either lack or are unable to leverage claim knowledge and select more \textit{refute} labels, both with CoT and SMART reasoning (the \%drop is higher for the smaller models).  However, this incorrect performance boost is worse with CoT than SMART (the \%drop is higher for CoT than SMART).  It is also worse for SFBench's Material Science knowledge than SciFact's Biological Science knowledge (the \%drop is higher SFBench than SciFact, Llama3 8b model on SciFact is an exception where the model does not generate most rationales). Web search based on a claim is noisy; up to half the RAG \textit{refute} labels are based on not finding the appropriate evidence.  Since RAG performance is lower than CoT, we provide \% SMART improvement over CoT in the last column (\textbf{SMART-PK improves performance up to 20\%.}) We discuss these results in \cref{sec:results}.   
 }
\label{apptab:model-cot-smart-performance}
%\vspace{-3mm}
\end{table*}

\begin{table*}[t]
 \centering
 \resizebox{.99\textwidth}{!}{
  \begin{tabular}{l|l|ccr||ccr||cc|}
 \hline
  &         &\multicolumn{3}{|c|}{Single Premise SMART} & \multicolumn{3}{|c|}{Contrastive Premises (SMART-PK)}  &\multicolumn{2}{|c|}{\%Contrast. Improv.} \\
Data & Model & Before RF&  After RF & \textbf{\%Drop$\downarrow$}& Before RF & After RF &\textbf{\%Drop$\downarrow$}   &Before RF & After RF\\
\hline
&Qwen3 1.7b         &.550$\pm$.005 &.330$\pm$.005 &22.0 &.535$\pm$.023 &.274$\pm$.025 &26.1 &-1.5 & -5.6\\
&Qwen3 8b           &.566$\pm$.005 &.392$\pm$.005 &17.4 &.568$\pm$.009 &.435$\pm$.016 &13.3 & 0.2 & 4.3\\
&Qwen3 32b          &.567$\pm$.012 &.508$\pm$.005 & 5.9 &.620$\pm$.019 &.569$\pm$.032 & 5.1 & 5.3 & 6.1\\
SFBench&Llama 8b    &.611$\pm$.016 &.442$\pm$.005 &16.9 &.579$\pm$.004 &.447$\pm$.010 &13.2 &-3.2 & 0.5\\
&Llama 70b          &.640$\pm$.006 &.492$\pm$.005 &14.8 &.542$\pm$.023 &.536$\pm$.030 & 0.6 &-9.8 & 4.4\\
&GPT-5.4-mini       &.760$\pm$.008 &.659$\pm$.005 & 6.1 &.750$\pm$.026 &.702$\pm$.004 & 4.8 &-1.0 & 0.3\\
&Cla. Haiku 4.5     &.652$\pm$.010 &.533$\pm$.005 &11.9 &.653$\pm$.012 &.583$\pm$.004 & 7.0 &-0.1 & 5.0\\
\hline
&Qwen3 1.7b         &.626$\pm$.010 &.432$\pm$.005 &19.4 &.646$\pm$.019 &.448$\pm$.044 &19.8 &2.0  &1.6\\
&Qwen3 8b           &.714$\pm$.002 &.599$\pm$.005 &11.5 &.722$\pm$.014 &.641$\pm$.007 & 8.1 &0.8  &4.2\\
SciFact \&&Qwen3 32b&.788$\pm$.005 &.675$\pm$.005 &11.3 &.731$\pm$.014 &.696$\pm$.018 & 3.5 &-5.7 &2.1\\
SciFact-&Llama 8b   &.650$\pm$.004 &.547$\pm$.005 &10.3 &.690$\pm$.025 &.595$\pm$.015 & 9.5 &-5.5 &4.8\\
Open&Llama 70b      &.658$\pm$.005 &.620$\pm$.005 & 3.8 &.665$\pm$.012 &.654$\pm$.014 & 1.1 &-0.4 &3.4\\
&GPT-5.4-mini       &.804$\pm$.006 &.743$\pm$.005 & 6.1 &.798$\pm$.006 &.765$\pm$.008 & 3.3 &-0.6 &2.2\\
&Cla. Haiku4.5      &.773$\pm$.008 &.712$\pm$.005 & 6.1 &.782$\pm$.014 &.749$\pm$.017 & 3.3 & 0.9 & 3.7\\
%\hline 
%&Qwen3 1.7b           &.563$\pm$.000 &.583$\pm$.014 &2.0 &.545$\pm$.010 &.606$\pm$.008 & 6.1\\
%&Qwen3 8b             &.613$\pm$.000 &.635$\pm$.005 &2.2 &.618$\pm$.082 &.630$\pm$.002 & 1.2\\
%&Qwen3 32b            &.652$\pm$.001 &.662$\pm$.010 &1.0 &.707$\pm$.083 &.645$\pm$.005 & 6.2\\
%AVeriTeC&Llama3.1 8b  &.541$\pm$.000 &.640$\pm$.014 &9.9 &.618$\pm$.092 &.611$\pm$.019 & -0.7\\
%&Llama3.1 70b         &.742$\pm$.008 &.726$\pm$.004 &-1.6 &.623$\pm$.057 &.732$\pm$.008 & 10.9\\
%&GPT-5.4-mini         &.800$\pm$.004 &.784$\pm$.026 &-1.6 &.617$\pm$.061 &.800$\pm$.035 & 18.3\\
%&Cla. Haiku 4.5       &.775$\pm$.002 &.784$\pm$.002 &0.9 &.577$\pm$.037 &.781$\pm$.003 & 20.4\\14 10
\hline
\end{tabular}
}
 \caption{Comparing the balanced accuracy of using Single Vs. Contrastive premises with the SMART framework.  We examine performance before and after applying the \textbf{Refute Filter} (\textit{refute} labels that are not adequately supported are changed to 'unknown', essentially dropping them from contributing to balanced accuracy). Single premises often lack necessary claim knowledge that  models select \textit{refute} labels without adequately supporting them. Correcting for this with the \textbf{Refute Filter} shows that contrastive knowledge improves on average across models and datasets by 3\%. The smallest model Qwen3 1.7b is worse at obtaining contrastive knowledge, but leads to better performance with single premises. We discuss these results in \cref{sec:results}.   
 }
\label{apptab:model-single-constrastive-performance}
%\vspace{-3mm}
\end{table*}

\begin{table*}[t]
 \centering
 \resizebox{.99\textwidth}{!}{
  \begin{tabular}{|l|l|c||cc||cc|}
 \hline
Data & Model &  SMART-PK & SMART-WS & \textbf{\%Imp.}$\uparrow$  & X-SMART-PK &\textbf{\%Imp.}$\uparrow$  \\
\hline
&Qwen3 1.7b           &.535$\pm$.023 &.544$\pm$.020 & 0.9  &.620$\pm$.015 & 8.5 \\
&Qwen3 8b             &.568$\pm$.009 &.598$\pm$.014 & 3.0  &.621$\pm$.013 & 5.3 \\
&Qwen3 32b            &.620$\pm$.019 &.611$\pm$.019 & -0.9 &.645$\pm$.021 & 2.5 \\
SFBench&Llama3.1 8b   &.579$\pm$.004 &.545$\pm$.029 & -3.4 &.578$\pm$.019 & 0.0 \\
&Llama3.1 70b         &.542$\pm$.023 &.610$\pm$.007 & 6.8  &.598$\pm$.019 & 5.6 \\
&GPT-5.4-mini         &.750$\pm$.026 &.736$\pm$.015 & -2.4 &.750$\pm$.031 & -   \\
&Cla. Haiku 4.5       &.653$\pm$.012 &.658$\pm$.035 & 0.5  &.677$\pm$.000 & 2.4 \\
\hline
&Qwen3 1.7b           &.646$\pm$.019 &.632$\pm$.015 & -1.4  &.692$\pm$.010 &4.6 \\
&Qwen3 8b             &.722$\pm$.014 &.732$\pm$.009 & 1.0  &.722$\pm$.014 &-    \\
SciFact \& &Qwen3 32b &.731$\pm$.014 &.734$\pm$.027 & 0.3  &.873$\pm$.010 &14.1 \\
SciFact-&Llama3.1 8b  &.690$\pm$.025 &.640$\pm$.013 & -5.0 &.727$\pm$.014 &3.7  \\
Open&Llama3.1 70b     &.665$\pm$.012 &.682$\pm$.012 & 1.7  &.796$\pm$.012 &13.1 \\
&GPT-5.4-mini         &.798$\pm$.006 &.788$\pm$.014 & -1.0 &.866$\pm$.010 &6.5  \\
&Cla. Haiku4.5        &.782$\pm$.014 &.784$\pm$.004 & 0.2 &.809$\pm$.012 &8.4    \\
\hline
\end{tabular}
}
 \caption{Comparing performance (BAccuracy) of SMART-WS and X-SMART-PK with SMART-PK. X-SMART-PK where models use premises from a better generating model: Qwen3 8b for SciFact and GPT-5.4-mini for SFBench.  Since SMART-PK does not lead to as much of an artificial performance boost, we compare results without applying the \textbf{Refute Filter}.  Performance on SciFact collection of claims improves by a 6.5-14.1\%; and up to 8.5\% for SFBench claims.  We discuss these results and choice of curation model in \cref{sec:results}.     
 }
\label{apptab:cross-model-results}
%\vspace{-3mm}
\end{table*}

\begin{table*}[t]
 \centering
 \resizebox{.99\textwidth}{!}{
  \begin{tabular}{l|l|c||cccc||r}
 \hline

Data & Model & SMART-PK& MajVot&AggJ&CriJ&SuppJ&\textbf{\%Improv.$\uparrow$}    \\

\hline
&Qwen3 1.7b           &.535$\pm$.023  &.536$\pm$.022 &.540$\pm$.009 &.520$\pm$.005 &.528$\pm$.010 &-0.1\\
&Qwen3 8b             &.568$\pm$.009  &.522$\pm$.014 &.558$\pm$.007 &.543$\pm$.016 &.586$\pm$.021 &4.6 \\
&Qwen3 32b            &.620$\pm$.019  &.530$\pm$.008 &.606$\pm$.019 &.575$\pm$.025 &.594$\pm$.019 &9.0 \\
SFBench&Llama3.1 8b   &.579$\pm$.004  &.464$\pm$.014 &.534$\pm$.027 &.529$\pm$.014 &.568$\pm$.022 &11.5 \\
&Llama3.1 70b         &.542$\pm$.023  &.506$\pm$.042 &.552$\pm$.015 &.553$\pm$.007 &.523$\pm$.005 &3.6\\
&GPT-5.4-mini         &.750$\pm$.026  &.528$\pm$.011 &.739$\pm$.012 &.646$\pm$.023 &.728$\pm$.030 &22.2\\
&Cla. Haiku 4.5       &.653$\pm$.012  &.593$\pm$.012 &.688$\pm$.008 &.685$\pm$.004 &.678$\pm$.039 &6.0\\
\hline
&Qwen3 1.7b           &.646$\pm$.019  &.613$\pm$.024 &.634$\pm$.017 &.631$\pm$.020 &.637$\pm$.008 & 3.3\\
&Qwen3 8b             &.722$\pm$.015  &.704$\pm$.003 &.722$\pm$.014 &.685$\pm$.011 &.725$\pm$.011 & 1.8 \\
SciFact \&&Qwen3 32b  &.731$\pm$.014  &.674$\pm$.019 &.730$\pm$.016 &.686$\pm$.003 &.692$\pm$.018 & 5.7 \\
SciFact-&Llama3.1 8b  &.690$\pm$.025  &.631$\pm$.015 &.691$\pm$.003 &.654$\pm$.020 &.631$\pm$.011 & 5.9\\
Open&Llama3.1 70b     &.665$\pm$.012  &.607$\pm$.010 &.658$\pm$.015 &.650$\pm$.017 &.632$\pm$.002 & 5.3\\
&GPT-5.4-mini         &.798$\pm$.006  &.720$\pm$.007 &.795$\pm$.002 &.757$\pm$.010 &.796$\pm$.012 & 7.8\\
&Cla. Haiku 4.5       &.782$\pm$.014  &.723$\pm$.022 &.788$\pm$.001 &.778$\pm$.016 &.733$\pm$.021 & 5.9\\
%\hline
%&Qwen3 1.7b           &.583$\pm$.014 &.598$\pm$.010&-1.5& .564$\pm$.026 &.586$\pm$.020 &.593$\pm$.008 &.596$\pm$.023 &1.9\\
%&Qwen3 8b             &.635$\pm$.005 &.628$\pm$.029 &0.7& .643$\pm$.010 &.640$\pm$.002 &.610$\pm$.023 &.654$\pm$.013 & -1.9\\
%&Qwen3 32b            &.677$\pm$.003 &.677$\pm$.003 &0.0& .671$\pm$.028 &.703$\pm$.022 &.618$\pm$.024 &.659$\pm$.019 & 0.6\\
%AVeriTeC&Llama3.1 8b  &.640$\pm$.014 &.624$\pm$.011 &1.6& .553$\pm$.014 &.609$\pm$.008 &.567$\pm$.008 &.577$\pm$.016 & 8.7\\
%&Llama3.1 70b         &.726$\pm$.004 &.749$\pm$.003&-2.3& .651$\pm$.018 &.729$\pm$.005 &.735$\pm$.020 &.734$\pm$.010 & 7.5\\
%&GPT-5.4-mini         &.784$\pm$.026 &.772$\pm$.012 &1.2& .698$\pm$.004 &.771$\pm$.025 &.722$\pm$.008 &.784$\pm$.012 & 8.6\\
%&Cla. Haiku 4.5       &.784$\pm$.002 &.774$\pm$.002 &1.0& .728$\pm$.017 &.774$\pm$.005 &.751$\pm$.010 &.755$\pm$.008& 5.6\\
 \hline
\end{tabular}
}
 \caption{Comparing the balanced accuracy of SMART-PK (with Reasoning Judge as meta-reasoner) with 4 other meta-reasoners. We show the advantage of using the Reasoning Judge over Majority Voting in the \%Improv. column.   
 }
\label{apptab:metareason-options}
%\vspace{-3mm}
\end{table*}

\begin{table*}[t]
 \centering
 \resizebox{.99\textwidth}{!}{
  \begin{tabular}{l|l|c||cr||cr||cr}
 \hline

Data & Model & SMART-NV&SMART-PK&\textbf{\%Improv$\uparrow$}& SMART-WS &\textbf{\%Improv$\uparrow$} &X-SMART-PK&\textbf{\%Improv$\uparrow$} \\

\hline
&Qwen3 1.7b          &.520$\pm$.005 &.535$\pm$.023 &1.5 &.544$\pm$.020 & 2.4 &.620$\pm$.015 &10.0 \\
&Qwen3 8b            &.542$\pm$.014 &.568$\pm$.009 &2.6 &.598$\pm$.014 & 5.6 &.621$\pm$.013 & 7.9 \\
&Qwen3 32b           &.564$\pm$.014 &.620$\pm$.019 &4.9 &.611$\pm$.019 & 4.7 &.645$\pm$.021 & 8.1 \\
SFBench&Llama3.1 8b  &.572$\pm$.016 &.579$\pm$.004 &0.7 &.545$\pm$.029 &-2.7 &.578$\pm$.019 & 0.6 \\
&Llama3.1 70b        &.518$\pm$.004 &.542$\pm$.023 &2.4 &.610$\pm$.007 & 9.2 &.598$\pm$.019 & 8.0 \\
&GPT-5.4-mini        &.753$\pm$.008 &.750$\pm$.026 &-0.3&.736$\pm$.015 &-1.7 &.750$\pm$.031 &-0.3 \\
&Cla. Haiku 4.5      &.650$\pm$.015 &.653$\pm$.012 &0.3 &.658$\pm$.035 & 0.8 &.677$\pm$.000 & 2.7\\
\hline
&Qwen3 1.7b          &.636$\pm$.012 &.646$\pm$.019 &1.0 &.632$\pm$.015 &-0.4 &.692$\pm$.010 & 5.6 \\
&Qwen3 8b            &.697$\pm$.002 &.722$\pm$.015 &2.5 &.732$\pm$.009 & 3.5 &.722$\pm$.014 & 2.5\\
SciFact \&&Qwen3 32b &.684$\pm$.011 &.731$\pm$.014 &4.7 &.734$\pm$.027 & 5.0 &.873$\pm$.010 &18.9 \\
SciFact-&Llama3.1 8b &.697$\pm$.003 &.690$\pm$.025 &-0.7&.640$\pm$.013 &-5.7 &.727$\pm$.014 & 3.0\\
Open&Llama3.1 70b    &.645$\pm$.018 &.665$\pm$.012 &2.0 &.682$\pm$.012 & 3.7 &.796$\pm$.012 &15.1 \\
&GPT-5.4-mini        &.771$\pm$.005 &.798$\pm$.006 &2.7 &.788$\pm$.014 & 1.7 &.866$\pm$.010 & 9.5 \\
&Cla. Haiku 4.5      &.765$\pm$.010 &.782$\pm$.014 &1.7 &.784$\pm$.004 & 1.9 &.809$\pm$.012 & 4.4 \\
 \hline
\end{tabular}
}
 \caption{Comparing the balanced accuracy of SMART without verification (SMART-NV) with self-verification (SMART-PK), external verification with websearch (SMART-WS) and cross-verification (X-SMART-PK). Improvement with each method is shown in \%Improve (the difference in performance between SMART-NV and the method).  Verification improves performance on average across models and datasets by by 2-7\%.  Grounding premises with real-world knowledge provided by websearch improves on average by 4-5\% with some exceptions.   
 }
\label{apptab:verification-options}
%\vspace{-3mm}
\end{table*}

\begin{table*}[t]
 \centering
 \resizebox{.99\textwidth}{!}{
  \begin{tabular}{l|l|c|cccc|cccc|c}
 \hline

 & &Model &\multicolumn{4}{|c|}{SMART-PK Inference } & \multicolumn{4}{|c}{SMART-WS Inference }  &Premise\\
Data & Model & Calls& KTok & \%Fback  &Sup. & Opp. & KTok& \%Fback &Sup. & Opp. &Quality \\
\hline
&Qwen3 1.7b          &5 &11&42&1&0 &11&70&.5&0 & NA\\
&Qwen3 8b            &11 &12&25&1.5&1.2 &11&83&.3&.1& NA\\
&Qwen3 32b           &19 &13&3&4.1&1.9 &12&23&2.0&.73& NA\\
SFBench&Llama 8b     &15 &20&4&2.7&1.6 &16&40&1.2&.7& NA\\
&Llama 70b           &18 &14&0&5&1.9 &11&63&.6&.2& NA\\
&GPT-5.4-mini        &19 &13&4&2.5&2 &12&10&2.0&1.5& NA\\
&Cla. Haiku 4.5      &20 &14&0&3.8&3.9 &12&12&2.3&1.6& NA\\
\hline
&Qwen3 1.7b          &4 &11&49&.7&.1 &11&67&.4&.1 &.41 \\
&Qwen3 8b            &11 &12&19&1.7&1.3 &12&42&1.0&.8&.36\\
SciFact \&&Qwen3 32b &18 &13&3&3.6&2.3 &12&.25&1.8&1.3&.36\\
SciFact-&Llama 8b    &14 &13&2&3.3&1.5 &13&14&2.6&1.2&.59\\
Open&Llama 70b       &17 &14&1&5.1&1.1 &11&47&1.7&.4&.32\\
 
&GPT-5.4-mini        &16 &12&8&2.3&1.6 &12&16&2.0&1.3&.28\\
&Cla. Haiku4.5       &20 &14&1&5.6&2.9 &13&4&3.6&2.2&.34\\
%\hline 
%&Qwen3 1.7b          &4 &11&59&.3&.3 &11&55&.13&.4 &.18\\
%&Qwen3 8b            &8 &11&48&.3&1.1 &11&66&.2&.7&.16\\
%&Qwen3 32b           &17 &13&9&1.4&2.8 &12&30&.7&1.8&.24\\
%AVeriTeC&Llama3.1 8b &13 &12&11&3.0&3.0 &12&30&1.2&1.0&.34\\
%&Llama3.1 70b        &14 &12&8&2.0&1.6 &11&51&.6&.6&.22\\
%&GPT-5.4-mini        &17 &12&10&1.3&2.1 &12&16&1.3&1.9&.31\\
%&Cla. Haiku 4.5      &18 &13&6&1.5&3.3 &12&22&1&2.2&.23\\
\hline
\end{tabular}
}
 \caption{Comparing inference cost for SMART-PK and SMART-WS in model calls and number of input $+$ output tokens measured in K tokens  (KTok). We report \% fallbacks (Fback) to the direct reasoning prompt when there are no curated premises, the quality of curated premises (Qual) and their average numbers per stance (Sup. \& Opp.).  Both methods require the same \# calls since we verify before obtaining stance; these do not include 10 calls that iteratively obtain premises as a single call could potentially be used to obtain all premises for  GPT-5.4-mini and Claude-Haiku-4.5. All values are per claim averages. We report \textbf{NA-Not applicable} as the premise quality (Qual) for SFbench claims due to not having textual evidence. Average premise quality by coincidence is the same for both methods, so we report in a single column.
 }
\label{apptab:inference-costs}
\vspace{-3mm}
\end{table*}

\begin{figure*}[t]
    \centering
     
    \begin{subfigure}[b]{0.32\textwidth}
    \includegraphics[trim=0 0.3in 0 0in,  width=1 \columnwidth]
    {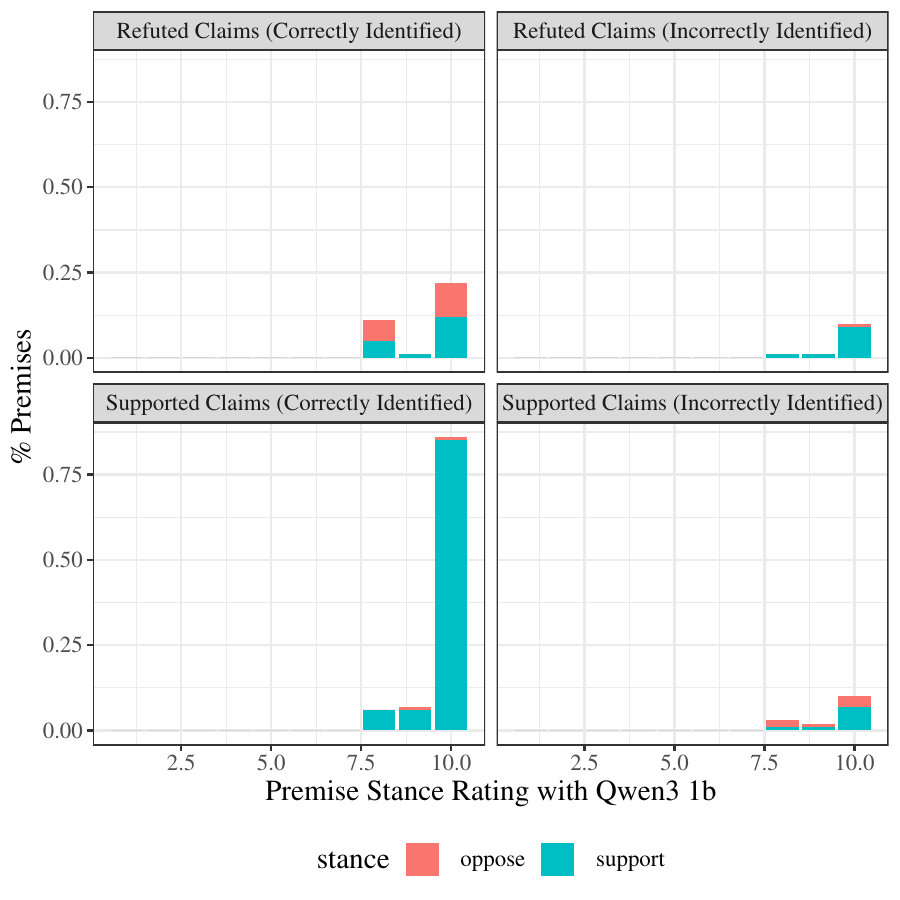} 
    \caption{ Qwen3 1.7b curated premises. } 
	\label{fig:stance-alignment-qwen1b}
 
\end{subfigure}
 ~        
    %\hfill
    \begin{subfigure}[b]{0.32\textwidth}
    \centering     
    \includegraphics[trim=0 0.3in 0 0in,  width=1 \columnwidth]
    {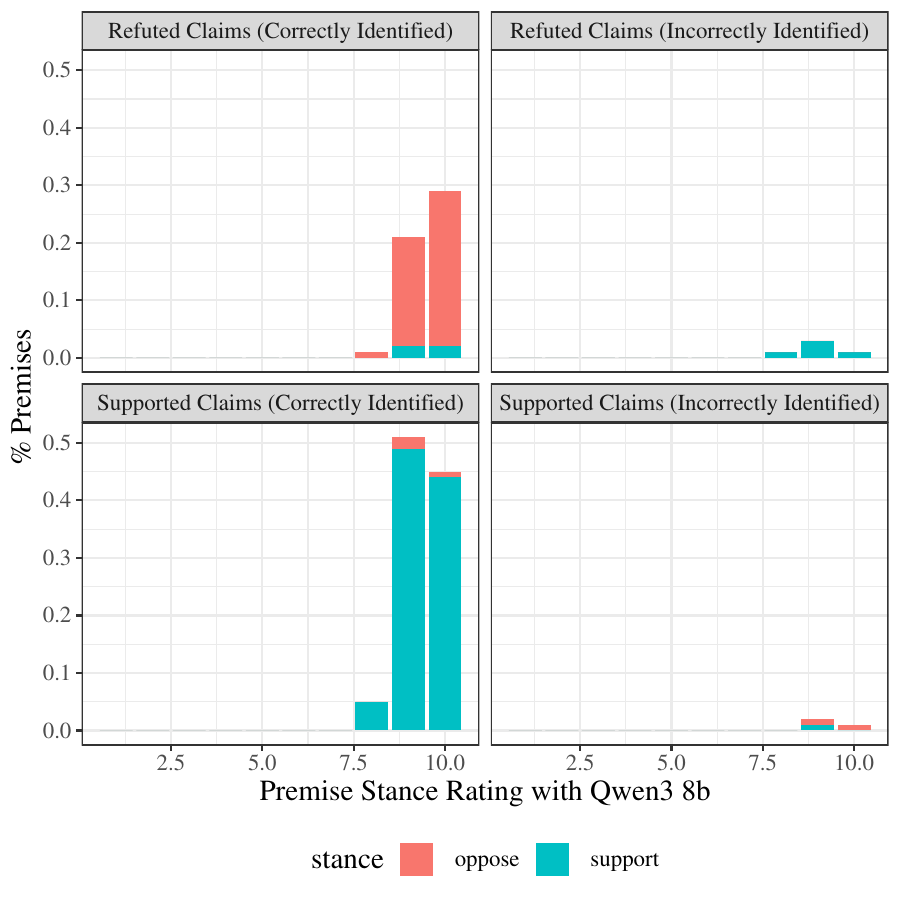} 
   \caption{Qwen3 8b curated premises.   } 
	\label{fig:stance-alignment-qwen8b}
 
\end{subfigure}
 ~        
    %\hfill
    \begin{subfigure}[b]{0.32\textwidth}    \centering     
    \includegraphics[trim=0 0.3in 0 0in,  width=1 \columnwidth]
    {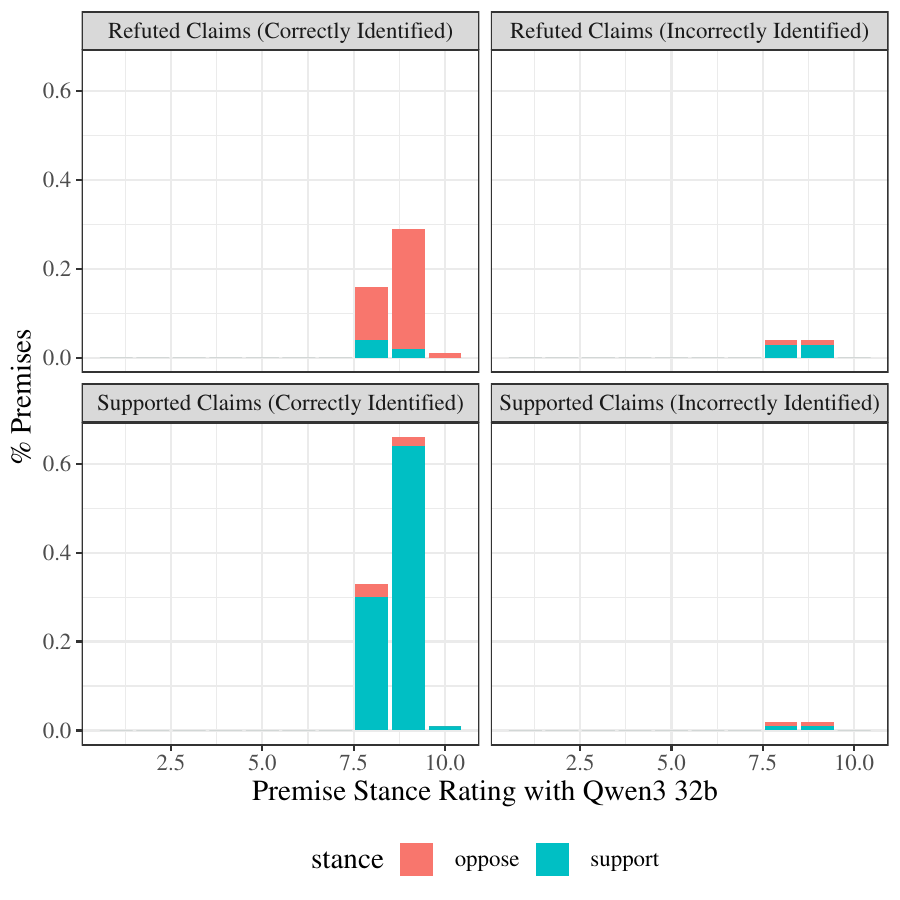} 
    \caption{ Qwen3 32b  curated premises.   } 
	\label{fig:stance-alignment-qwen32b}
 \end{subfigure}
 \caption{Distribution of curated premise stances, their alignment with the claim and their confidence ratings for Qwen3 models.  Correct identification of \textit{refuted} claims requires opposing premises and \textit{supported} claims require supporting premises.  \textbf{Qwen3 1.7b curation leads to few and noisy premises compared to Qwen8b.  Qwen3 32 generates too many supporting premises increasing meta-reasoning difficulty. Supporting and opposing premises are required for supporting claims of \textit{support} and \textit{refute} labels respectively.}}
 \label{fig:stance-alignment-qwen}
\end{figure*}

\begin{figure*}[t]
    \centering     
    \includegraphics[trim=0 0in 0 0in,  width=.98 \textwidth]
    {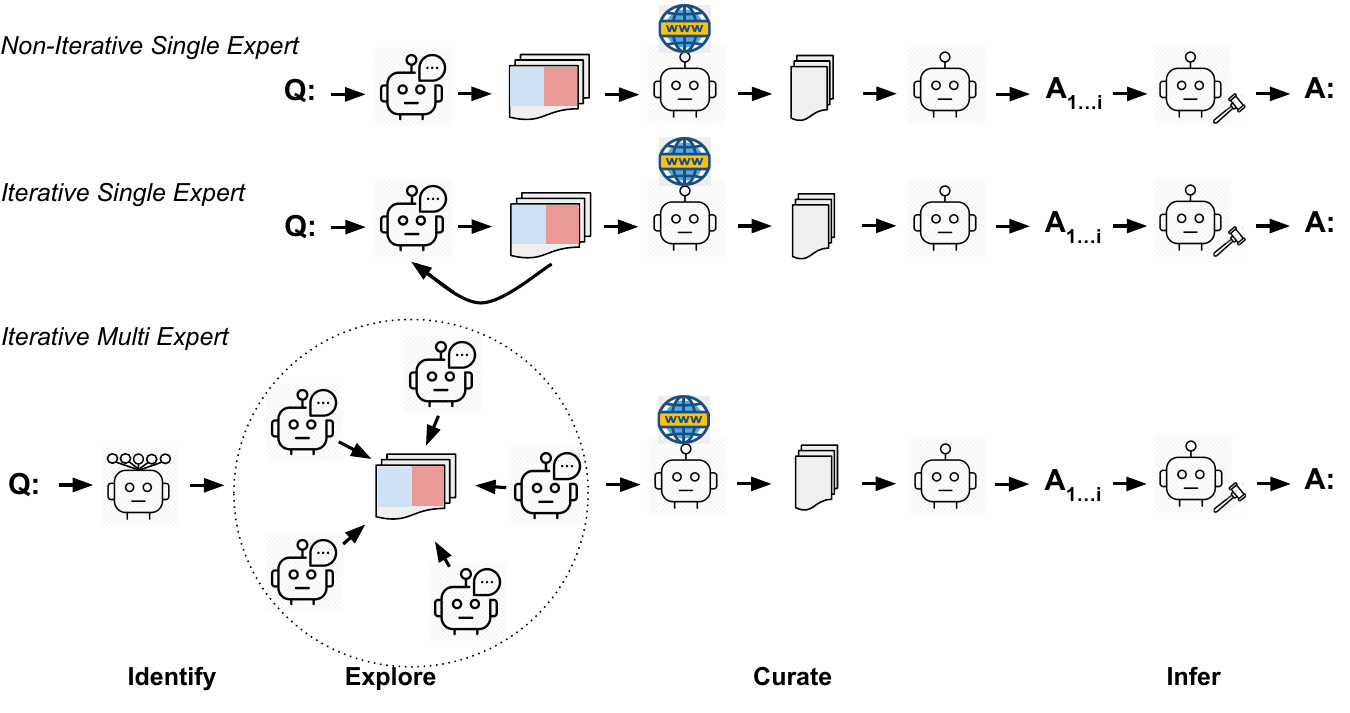} 
    \caption{We compare various configurations we  used for generating premises. The performance of these different configurations is compared in \cref{fig:priming-types}.
    } %
	\label{fig:configurations}
 \vspace{-5mm}
\end{figure*}

 \begin{table*}[t]
 \centering
\resizebox{.98\textwidth}{!}{
 \begin{tabular}[trim=0 .2in 0 0,  width=.95\linewidth]{|p{0.14\linewidth}|p{0.85\linewidth} |}
 \hline
Models& Contrastive premise pairs (separated by an \&; ... replaces repeating text from premise \#1) \\ 
\hline
\multicolumn{2}{|l|}{\textbf{\cref{fig:fig1} Example Claim: Genetic deletion of JAM-A decreases gut permeability.}} \\
\hline
Qwen3 1b & \textbf{Premises are few and superficial.} \\
&Genetic deletion of JAM-A occurs in a specific cellular context. \& Genetic deletion of JAM-A does not occur in a specific cellular context. \\
&Genetic deletion of JAM-A is associated with a specific cellular location. \& Genetic deletion of JAM-A is not associated with a specific cellular location.\\
\hline 
Qwen3 8b & \textbf{Premises are atomic and easy to validate.} \\
&JAM-A is involved in maintaining gut barrier integrity. \& JAM-A is not ... integrity.\\
& JAM-A promotes the formation of tight junctions in intestinal epithelial cells. \&  JAM-A inhibits the formation of tight junctions in intestinal epithelial cells.\\
& Genetic deletion of JAM-A reduces the expression of tight junction proteins. \& Genetic deletion of JAM-A increases the expression of tight junction proteins. \\
&JAM-A is a component of the intestinal epithelial cell membrane. \& JAM-A is not ...\\ &Genetic deletion of JAM-A disrupts the structural integrity of the intestinal epithelial barrier. \& Genetic deletion of JAM-A enhances the structural  ... epithelial barrier.\\
&Genetic deletion of JAM-A decreases the expression of claudin proteins. \& Genetic deletion of JAM-A increases the expression of claudin proteins.\\
& Genetic deletion of JAM-A decreases the expression of occludin proteins.\& Genetic deletion of JAM-A increases the expression of occludin proteins.\\
&Genetic deletion of JAM-A increases the paracellular transport of molecules across the intestinal epithelium. \& Genetic deletion of JAM-A decreases ...  intestinal epithelium.\\
\hline 
Qwen3 32b & \textbf{Premises are more nuanced, as result these are less atomic and difficult to validate.}\\
& JAM-A is normally expressed in gut epithelial cells. \& JAM-A is not expressed ...  \\
& The intestinal barrier is intact.\& The intestinal barrier is compromised.\\ 
&JAM-A deletion leads to reduced paracellular transport. \& JAM-A deletion does not ...\\
&JAM-A is required for tight junction integrity. \& JAM-A is not required ...  \\
&JAM-A deletion increases paracellular permeability. \& JAM-A deletion decreases ...\\ &JAM-A deletion enhances intestinal barrier function. \& JAM-A deletion impairs ... \\
&JAM-A deletion reduces intestinal paracellular permeability. \& JAM-A deletion increa ...\\
&JAM-A is essential for maintaining gut epithelial barrier integrity. \& JAM-A is not ...\\ 
&JAM-A deletion reduces the passage of molecules through the gut epithelium. \& JAM-A deletion increases the passage of molecules through the gut epithelium.\\
&JAM-A deletion is associated with decreased paracellular transport of luminal antigens. \& JAM-A deletion is associated with increased paracellular transport of luminal antigens.\\
\hline
GPT-5.4-mini& \textbf{Premises are more scientific \& salient; not easy for smaller models to validate.}\\
&JAM-A has a barrier-stabilizing role in intestinal epithelial tight junctions. \& JAM-A has a barrier-weakening role in intestinal epithelial tight junctions.\\
& Claudin-1 abundance is elevated in the intestinal epithelium. \& Claudin-1 abundance is not elevated in the intestinal epithelium.\\
&The JAM-A deletion is restricted to intestinal epithelial cells. \& The JAM-A deletion is not restricted to intestinal epithelial cells.\\
&The intestinal mucosa is inflamed. \& The intestinal mucosa is not inflamed.\\ 
&Occludin abundance is elevated in the intestinal epithelium. \& Occludin ... is not ...\\
&Intestinal epithelial apoptosis is increased. \& Intestinal epithelial apoptosis is not ...\\ 
&Tricellular tight junction sealing is compromised in the intestinal epithelium. \& Tricellular tight junction sealing is intact in the intestinal epithelium.\\
&JAM-A is localized at intestinal epithelial tight junctions. \& JAM-A is not localized ...\\
&Claudin-2 abundance is reduced in the intestinal epithelium. \& Claudin-2 abundance is not reduced in the intestinal epithelium.\\
&Myosin light chain phosphorylation is reduced in the intestinal epithelium. \& Myosin light chain phosphorylation is not reduced in the intestinal epithelium.\\

\hline
\end{tabular}
}
 \caption{Example premises with different Qwen3 and GPT-5.4-mini models. See \cref{apptab:premise-examples-more} for other models.}
\label{apptab:premise-examples}
\end{table*}

 \begin{table*}[t]
 \centering
\resizebox{.98\textwidth}{!}{
 \begin{tabular}[trim=0 .2in 0 0,  width=.95\linewidth]{|p{0.14\linewidth}|p{0.85\linewidth} |}
 \hline
Models& Contrastive premise pairs (separated by an \&; ... replaces repeating text from premise \#1) \\ 
\hline
\multicolumn{2}{|l|}{\textbf{\cref{fig:fig1} Example Claim: Genetic deletion of JAM-A decreases gut permeability.}} \\
\hline
Llama3.1 8b & \textbf{Premises are limited to the context and do not explore new entities and concepts.} \\
&JAM-A is expressed in the epithelial cells of the gut \& JAM-A is not expressed ...\\ &Gut permeability is not affected by JAM-A deletion \& Gut permeability is increased by ...\\
&The mechanism of JAM-A's effect on gut permeability is well understood \& The mechanism of JAM-A's effect on gut permeability is not well understood\\
&Gut permeability decreases when JAM-A is not present. \& Gut permeability does not ...\\
&JAM-A deletion leads to a significant decrease in gut permeability \& JAM-A deletion has no effect on gut permeability\\
&JAM-A deletion results in the disruption of tight junctions in the gut epithelial cells \& JAM-A deletion does not affect the structure of tight junctions in the gut epithelial cells\\
&JAM-A deletion leads to a decrease in gut permeability is a well-established scientific fact \& JAM-A deletion leads to an increase in gut permeability\\
&JAM-A is a key regulator of tight junctions in the gut epithelial cells \& JAM-A is not involved in regulating tight junctions in the gut epithelial cells\\
\hline 
Llama3.1 70b & \textbf{Premises are simple and atomic but most premises support the claim.}  \\
&JAM-A is involved in the regulation of tight junctions in the gut epithelium \& JAM-A is not involved in the regulation of tight junctions in the gut epithelium\\
&JAM-A is a crucial component of the intestinal epithelial barrier \& JAM-A is not a cruc ...\\
&The deletion of JAM-A does not affect the expression of other tight junction proteins in the gut epithelium \& The deletion of JAM-A affects the expression ...\\
&The gut epithelium of genetically deleted JAM-A mice has increased levels of pro-inflammatory cytokines.\& The gut epithelium ... deleted JAM-A mice does not...\\ 
&JAM-A plays a role in the maintenance of the gut epithelial barrier's physical structure \& JAM-A does not play a role in the maintenance of the gut epithelial barrier's ...\\ 
&JAM-A is involved in the regulation of gut epithelial cell apoptosis \& JAM-A is not  ...\\
&The genetically deleted JAM-A mice have reduced gut motility \& The genetically deleted JAM-A mice do not have reduced gut motility\\
&JAM-A deletion reduces the expression of occludin in the gut epithelium \& JAM-A deletion does not affect the expression of occludin ...\\
\hline 
Claude-Haiku&\textbf{Premises are more scientific; not easy for smaller models to validate.} \\
-4.5&Tight junction protein complexes are intact and functional at the intestinal epithelial barrier. \& Tight junction protein complexes are disrupted or have compromised function ...\\
&JAM-A maintains adherens junction stability through interaction with VE-cadherin and catenin complexes. \& JAM-A does not contribute to adherens ...\\
&JAM-A regulates claudin-based tight junction strand formation and sealing function. \& JAM-A does not contribute to claudin-based tight junction strand formation or ...\\
&JAM-A is present and participates in leukocyte transepithelial migration regulation at the intestinal epithelial barrier. \& JAM-A is absent and cannot regulate leukocyte transepithelial migration, leading to increased immune cell infiltration and barrier destabilization.\\
&JAM-A is localized at tricellular junctions and forms sealing strands that prevent paracellular solute flux. \& JAM-A is absent from tricellular junctions and cannot form sealing strands, allowing increased paracellular solute permeability.\\
&JAM-A mediates homophilic interactions between adjacent epithelial cells that reinforce barrier cohesion and reduce paracellular permeability. \& JAM-A does not mediate homophilic cell-cell interactions or such interactions are non-essential for maintain ...\\
&JAM-A negatively regulates paracellular transport through size-selective barrier function at tricellular corners. \& JAM-A does not regulate paracellular transport selectivity or plays a minor role in size-dependent barrier permeability.\\
&JAM-A deletion results in compensatory upregulation of alternative tight junction proteins that maintain barrier function. \& JAM-A deletion does not trigger compensatory upregulation of alternative tight junction proteins, leaving barrier function compromised.\\
\hline 
\end{tabular}
}
 \caption{Example premises with Llama3 and Claude models.  See \cref{apptab:premise-examples} for other models.}
\label{apptab:premise-examples-more}
\end{table*}

 \begin{table*}[t]
 \centering
\resizebox{.98\textwidth}{!}{
 \begin{tabular}[trim=0 .2in 0 0,  width=.95\linewidth]{p{0.08\linewidth}|p{0.91\linewidth}   }
 
 \hline
Error & Issue Description \\ 
\hline
&\textbf{Claim:}Fluorine-containing additives in a liquid electrolyte enable Li-ion battery chemistries to cycle up to 10 V.\\
Type 2b&\textbf{CoT with Qwen3 8b:} \textit{refuted} label is \textbf{correct}.
\textbf{Rationale:} While fluorine-containing additives in electrolytes can improve the performance and stability of Li-ion batteries, there is no widely accepted evidence that they enable cycling up to 10 V. The typical voltage range for Li-ion batteries is around 3.0–4.3 V, and achieving 10 V would require significant technological advancements beyond current understanding and practice.\\
&\textbf{SMART with Qwen3 8b :} \textit{supported} label is \textbf{incorrect}.
\textbf{Rationale:} The claim is supported by the evidence that fluorine-containing additives increase voltage capacity and enhance electrochemical stability at high voltages. While there are opposing conditions suggesting potential degradation, the supporting conditions are more specific and credible, indicating that fluorine additives can enable Li-ion batteries to cycle up to 10 V.\\
\hline
&\textbf{Claim:} A battery with a lithium titanate anode, a lithium manganese oxide cathode, and an electrolyte consisting of 1M lithium hexafluorophosphate in dimethyl carbonate can pass cold crank at -10'C\\
Type 1&\textbf{CoT with Qwen3 8b:} Lithium titanate (Li4Ti5O12) anodes and lithium manganese oxide (LiMn2O4) cathodes are known for their good low-temperature performance. The electrolyte described, 1M lithium hexafluorophosphate in dimethyl carbonate, is commonly used in lithium-ion batteries and can function effectively at low temperatures, including -10°C. Therefore, the claim is supported by current scientific knowledge. \\
&\textbf{SMART with Qwen3 8b :} The claim is not supported because the opposing conditions indicate that the electrolyte loses ionic conductivity, has high viscosity, and the anode has low electrical conductivity at low temperatures, all of which hinder cold crank performance at -10°C.\\
Type 3&\textbf{CoT with Qwen3 32b:} Lithium titanate anodes and lithium manganese oxide cathodes are known for their stability and performance at low temperatures. Carbon cloth as a current collector is flexible and suitable for wearable applications. The freezing point of the electrolyte (-50°C) and the use of polypropylene encapsulation suggest a design capable of withstanding -30°C. While the specific integration on a finger joint is novel, the materials and properties described are plausible for the claim, though experimental validation is not provided.\\
&\textbf{SMART with Qwen3 32b :} While the carbon cloth, lithium titanate anode, and polypropylene encapsulation suggest potential for cold performance and flexibility, the opposing conditions indicate significant challenges: loss of adhesion, rigidity, ionic conductivity, and mechanical durability at -30°C would likely prevent the battery from maintaining structural integrity and function on a moving finger joint. The claim is not strongly supported due to these critical performance limitations.\\
Type 2a&\textbf{CoT with Llama3.1 70b:} \textit{infeasible} label is \textbf{correct}.
The claim is supported because the described battery components, including the lithium titanate anode, lithium manganese oxide cathode, and electrolyte with a freezing point of -50'C, are suitable for low-temperature operation, and the use of polypropylene encapsulation and carbon cloth substrates provides sufficient structural integrity and flexibility for mounting on a glove joint without interfering with mobility at -30'C.\\
&\textbf{SMART with Llama3.1 70b :} \textit{infeasible} label is \textbf{correct}.
The claim is not supported due to several opposing conditions that outweigh the supporting conditions, including the potential loss of mechanical strength of the carbon cloth substrate, low ionic conductivity of the lithium manganese oxide cathode, and potential cracking of the polypropylene encapsulation due to impact from finger movement, which could compromise the structural integrity and mobility of the battery at -30'C.\\
\hline 
\end{tabular}
}
 \caption{Errors seen in SFBench Data. With Qwen3 8b, 32b and Llama3.1 70b, there are more Type 2a \& 2b situations than Type 1 leading to SMART-PK's lower performance; SMART-WS improves upon some of the Type 2a \& 2b errors through external premise validation. See error descriptions in \cref{appsec:error-types} and \cref{apptab:scifact-examples} for SciFact errors.}
\label{apptab:sfbench-examples}
\end{table*}

 \begin{table*}[t]
 \centering
\resizebox{.98\textwidth}{!}{
 \begin{tabular}[trim=0 .2in 0 0,  width=.95\linewidth]{p{0.07\linewidth}|p{0.92\linewidth} }
 \hline
Error&Issue Description \\ 
\hline
&\textbf{Claim: } Chenodeosycholic acid treatment reduces whole-body energy expenditure.\\
Type 1&\textbf{CoT with Qwen3 32b:} \textit{feasible} label is \textbf{incorrect}. Chenodeoxycholic acid (CDCA) is a bile acid and FXR agonist, and FXR activation has been shown in experimental studies to suppress thermogenic pathways and lower whole-body energy expenditure. The evidence is mainly preclinical and context-dependent, so the support is good but not absolute.\\
&\textbf{SMART with Qwen3 32b :} \textit{infeasible} label is \textbf{correct}.
Chenodeoxycholic acid is a bile acid that can activate pathways such as TGR5, which promotes type 2 deiodinase activity in thermogenic tissue and tends to increase thermogenesis and whole-body energy expenditure. Given these opposing conditions, the claim that it reduces energy expenditure is not supported and is likely opposite to the expected biological effect.\\
&\textbf{Claim: } Macrolides have no protective effect against myocardial infarction.\\
Type 2a&\textbf{CoT with Qwen3 32b:} \textit{infeasible} label is \textbf{correct}. Current scientific evidence does not strongly support the claim that macrolides protect against myocardial infarction. While some studies have suggested potential anti-inflammatory and immunomodulatory effects of macrolides, these effects have not been conclusively linked to a protective role against myocardial infarction in large, well-controlled clinical trials.\\
&\textbf{SMART with Qwen3 32b :} \textit{feasible} label is \textbf{incorrect}. The claim that macrolides protect against myocardial infarction is supported by multiple plausible biological mechanisms, including anti-inflammatory, immunomodulatory, antioxidant, and anti-platelet effects, all of which are relevant to cardiovascular disease and myocardial infarction prevention. While there are no opposing conditions presented, the evidence is largely mechanistic and not based on strong clinical trial data showing direct protection against myocardial infarction, which limits the confidence to a high but not maximum level.\\
Type 3&\textbf{CoT with Llama3.1 70b:} \textit{feasible} label is \textbf{incorrect}. Current scientific knowledge suggests that macrolides, a class of antibiotics, have been found to have neutral or conflicting effects on cardiovascular outcomes, including myocardial infarction. While some studies have investigated potential associations, there is no conclusive evidence to support a protective effect of macrolides against myocardial infarction, leading to a decision of 'Y' with a moderate to high confidence rating.\\
&\textbf{SMART with Llama3.1 70b:} \textit{feasible} label is \textbf{incorrect}. The claim that macrolides protect against myocardial infarction is supported due to their anti-inflammatory properties and the conditions provided, such as the patient having an inflammatory condition responsive to macrolides and high levels of CRP and IL-6 in the blood, which are associated with cardiovascular disease. However, the rating is not a 10 due to the limited scope of the conditions and the lack of direct evidence linking macrolides to myocardial infarction prevention in all cases.\\
\hline 
\end{tabular}
}
 \caption{Errors seen in SciFact Data. With   Llama3.1 70b, GPT-5.4-mini and Claude-Haiku-4.5, there are a more Type 2a situations than Type 1 leading to SMART-PK's lower performance.  X-SMART-PK reduces some of the Type 3 and Type 2a errors leading to improved performance. See error descriptions in \cref{appsec:error-types} and \cref{apptab:scifact-examples} for SFBench errors.}
\label{apptab:scifact-examples}
\end{table*}

\newpage
\clearpage
\appendix

\end{document}